\documentclass[a4paper,fleqn]{cas-dc}

\usepackage[numbers]{natbib}
\usepackage{url}
\usepackage{xcolor}

\usepackage{hyperref}
\usepackage{multirow}
\usepackage{newfloat}
\usepackage{listings}
\usepackage{epsfig}
\usepackage{amsmath}
\usepackage{amssymb}
\usepackage{booktabs}
\usepackage{amsmath}
\usepackage{multirow}
\usepackage{float}
\usepackage{xcolor}
\usepackage{multirow}
\usepackage{rotating}
\usepackage[T1]{fontenc}
\usepackage{booktabs, makecell, tabularx}
\usepackage{pifont}
\newcommand{\cmark}{\ding{51}}%
\newcommand{\xmark}{\ding{55}}
\usepackage{graphicx}
\usepackage{comment}
\usepackage[export]{adjustbox}
\usepackage{algorithmic}
\usepackage{textcomp}
\usepackage{multirow,multicol, makecell, booktabs}

\usepackage{graphicx}
\usepackage{amsmath,amssymb}
\usepackage{booktabs, makecell, tabularx}
\usepackage{rotating}
\usepackage[export]{adjustbox}
\usepackage{booktabs}
\usepackage{subcaption}
\usepackage{graphicx}
\usepackage{pifont}

\def\tsc#1{\csdef{#1}{\textsc{\lowercase{#1}}\xspace}}
\tsc{WGM}
\tsc{QE}

\begin{document}
\let\WriteBookmarks\relax
\def\floatpagepagefraction{1}
\def\textpagefraction{.001}

\shorttitle{MambaCAFU for Medical Image Segmentation.}    

\shortauthors{Bougourzi Fares et al.}  

\title[mode = title]{MambaCAFU: Hybrid Multi-Scale and Multi-Attention Model with Mamba-Based Fusion for Medical Image Segmentation.}  
\author[1]{T-Mai Bui}
\ead{tbuihuynh001@ikasle.ehu.eus}
\author[1]{Fares Bougourzi}
\ead{faresbougourzi@gmail.com}
\author[1,2] {Fadi Dornaika}
\cormark[1]
\ead{fadi.dornaika@ehu.eus}
\author[3]{Vinh Truong Hoang}
\ead{vinh.th@ou.edu.vn}

\address[1]{University of the Basque Country UPV/EHU, San Sebastian, Spain}
\address[2]{IKERBASQUE, Basque Foundation for Science, Bilbao, Spain}
\address[3]{Ho Chi Minh City Open University, VietNam}

\tnotetext[1]{Corresponding Author} 

%


\begin{abstract}
In recent years, deep learning has demonstrated remarkable potential in achieving medical-expert-level performance for segmenting complex medical imaging tissues and tumors. However, current models face several limitations. Many of these approaches are task-specific, with performance varying significantly across different modalities and anatomical regions. Moreover, achieving an optimal trade-off between model complexity and performance remains an open challenge, especially in real-world clinical environments where both accuracy and efficiency are critical.
To address these limitations, we propose a novel hybrid medical imaging segmentation architecture. Our model features a three-branch encoder that integrates Convolutional Neural Networks (CNNs), Transformers, and a Mamba-based Attention Fusion (MAF) mechanism to leverage complementary strengths in capturing local, global, and long-range dependencies. A multi-scale attention-based CNN decoder is then employed to reconstruct fine-grained segmentation maps while preserving contextual consistency. Furthermore, we introduce a co-attention gate, which enhances feature selection by adaptively emphasizing relevant spatial and semantic information across different scales during both encoding and decoding phases. This mechanism facilitates improved feature interaction and cross-scale communication, leading to more precise and robust segmentation outcomes.
Extensive experiments across multiple benchmark datasets demonstrate that our approach consistently outperforms existing state-of-the-art methods in terms of both accuracy and generalization, while maintaining computational complexity comparable to that of average models. By effectively balancing efficiency and effectiveness, our architecture represents a practical and scalable solution for diverse medical imaging segmentation tasks. The source codes and trained models are available at \href{https://github.com/faresbougourzi/MambaCAFU}{MambaCAFU}.
\end{abstract}


\begin{keywords}
Mamba\sep Transformer\sep CNN\sep Co-Attention\sep Attention\sep Medical Image Segmentation. 
\end{keywords}

\maketitle

\section{Introduction}
Medical imaging plays a critical role in various clinical scenarios \cite{bougourzi2025recent}. The segmentation of tumors, abnormalities, and organs in medical images is a crucial step for disease diagnosis and treatment planning. This process involves distinguishing pixels belonging to organs or lesions in imaging modalities such as CT \cite{gu2021survey}, MRI \cite{lundervold2019overview}, and video endoscopy \cite{paderno2021videomics}.
With advancements in machine learning, deep learning models have demonstrated great potential in achieving expert-level performance in medical image segmentation. However, these models require large amounts of labeled data for training, which is often difficult to obtain. Moreover, their performance varies significantly across different tasks. To address these challenges, numerous deep learning models have been developed over the past decade \cite{bougourzi2025recent}.

A major breakthrough in medical image segmentation came with the development of the U-Net architecture \cite{bougourzi2024emb, 10.1007/978-3-319-24574-4_28}. U-Net is a convolutional neural network (CNN) designed in an encoder-decoder structure with a characteristic "U" shape, where encoder layers are connected to corresponding decoder layers via skip connections. The success of U-Net inspired numerous extensions, such as Attention U-Net \cite{oktay2018attention} and U-Net++ \cite{zhou_unet_2018}. These CNN-based architectures excel at learning local features due to their receptive field being limited to neighboring pixels. However, they struggle to model long-range dependencies effectively.

Transformers have revolutionized computer vision \cite{dosovitskiy2020vit, chen2021transunet} by capturing long-range dependencies through the self-attention (SA) mechanism. In medical image segmentation, they have been incorporated into various architectures, including pure Transformer-based models \cite{liu_swin_2021} and hybrid models that integrate CNNs with Transformers \cite{chen2021transunet, wang2022uctransnet}. However, self-attention suffers from quadratic computational complexity. Consequently, while Transformers can enhance performance, they also increase model size and computational demands.


To address these computational challenges, Mamba \cite{mamba} has been introduced, achieving state-of-the-art performance across multiple domains, including language, audio, and genomics. However, the application of Mamba in medical image segmentation is still in its early stages. Current architectures exhibit weak generalization across different segmentation tasks and imaging modalities, as will be demonstrated in the experimental section.

In this paper, we introduce a hybrid medical image segmentation model, MambaCAFU, which follows the U-Net structure. The encoder integrates deep features from three branches (PVT, Main, and ResNet) and incorporates the CoASMamba block. This block leverages multiple attention mechanisms to efficiently merge multi-level features, enhancing visual representation for improved segmentation performance. Furthermore, we introduce the CoAMamba module to refine and extract deeper features from two encoder layers in the bottleneck. Finally, the decoder is designed with our proposed DoubleLCoA block, which integrates Co-Attention and convolutional layers. The effectiveness of our proposed model is demonstrated across various medical imaging tasks and modalities. Our main contributions are:

\begin{itemize}

\item We propose MambaCAFU, a novel hybrid architecture that integrates Mamba, Transformers, and CNNs with an Efficient Multi-Scale Cross Attention mechanism for medical image segmentation.

\item Our three-branch hybrid encoder effectively fuses local CNN-based features and global Transformer-based features using our proposed Mamba-based Attention Fusion Module.

\item We introduce CoAMamba, a novel module designed to refine and extract deeper features in the bottleneck by leveraging multi-attention mechanisms.

\item Our decoder features the DoubleLCoA block, which combines consecutive Co-Attention Gates and convolutional layers to enhance segmentation performance.

\item Our approach achieves state-of-the-art performance across multiple medical image segmentation tasks and imaging modalities on six benchmark datasets, outperforming extensive SOTA methods.

\end{itemize}

This paper is organized as follows. Section \ref{S:2} reviews recent related works in medical image segmentation. Section \ref{S:3} describes the proposed approach. Section \ref{S:4} presents the datasets used to evaluate and compare our method with state-of-the-art (SOTA) approaches. Section \ref{S:5} details the evaluation metrics and training settings. Section \ref{S:6} reports the experimental results and comparison with SOTA approaches. Finally, Section \ref{S:7} concludes the paper.

\section{Related work}
\label{S:2}
Advancements in medical imaging segmentation architectures have evolved over the years, progressing from CNNs to Transformers, and more recently to Mamba-based approaches.

\noindent\textbf{Convolutional Neural Network}: Since the success of the U-Net architecture \cite{10.1007/978-3-319-24574-4_28}, numerous U-Net variants have been proposed, such as Attention U-Net \cite{oktay2018attention} and U-Net++ \cite{zhou_unet_2018}. The U-Net architecture features an encoder-decoder structure with skip connections that integrate features from the encoder into the decoder \cite{10.1007/978-3-319-24574-4_28}.
In the Attention U-Net architecture \cite{oktay2018attention}, Attention Gates (AGs) were introduced to refine the skip connections by emphasizing prominent feature locations. This mechanism enhances the integration of encoder features with their corresponding decoder counterparts, improving segmentation accuracy. Similarly, U-Net++ \cite{zhou_unet_2018}, proposed by Z. Zhou et al., incorporates convolutional bridge blocks that process encoder features during the skipping phase, ensuring a smoother flow of information from the encoder to the decoder. While convolutional neural network (CNN) blocks are highly effective in capturing local features, they are inherently limited in their ability to encode long-range dependencies.

\noindent\textbf{Transformer}: The success of Transformer models in natural language processing (NLP) has significantly influenced their adoption in various computer vision fields, including medical imaging segmentation (MIS). Numerous Transformer-based architectures have been proposed for MIS, primarily due to their ability to encode long-range dependencies effectively \cite{liu_swin_2021, chen2021transunet, wang2022uctransnet}. 
Various approaches have been explored to utilize Transformer
 architectures in MIS, including fully Transformer-based models \cite{liu_swin_2021} and hybrid CNN models \cite{chen2021transunet, wang2022uctransnet, bougourzi2024d}.

Despite the advantages of using Transformers in MIS, their application often comes with substantial computational costs, with these models typically exceeding 100 million trainable parameters \cite{bougourzi2024rethinking, rahman2024multi, chen2021transunet}. This is primarily due to the self-attention mechanism, the core component of Transformer architectures, which has a quadratic computational complexity with respect to the input sequence length. Additionally, training these models can pose convergence challenges. These limitations hinder the widespread adoption of Transformer-based architectures in medical imaging segmentation.

\noindent\textbf{Mamba}: 
As an alternative to heavy Transformer models, Mamba has been proposed as a parameter-efficient approach that replaces the transformer's self-attention with structured State Space Models (SSMs) \cite{gu2022efficiently}. SSMs are capable of capturing long-range dependencies with linear complexity \cite{gu2023mamba}. These models have been extended from NLP to the vision domain, where several modules have been proposed to adapt SSMs to the 2D structure of images, such as the Vim encoder \cite{zhu2024vision} and VSS Block \cite{Liu2024VMambaVS}. These adaptations have also been integrated into various architectures for medical imaging segmentation \cite{linguraru_swin-umamba_2024, wang2024mamba, U-Mamba, 10.1007/978-981-97-5128-0_27, ruan2024vmunetvisionmambaunet}.

In \cite{U-Mamba}, J. Ma et al. introduced U-Mamba, which features a hybrid CNN-Mamba encoder and a CNN decoder, connected via U-Net-like skip connections. The proposed Mamba block consists of two sequential components: two residual blocks followed by a custom Mamba block, which includes linear, convolutional, and SSM modules. Similarly, Mamba-UNet \cite{wang2024mamba} adapted the VSS Block \cite{Liu2024VMambaVS} to construct a hierarchical segmentation architecture, akin to Swin-UNet \cite{cao2022swin}.

Despite these efforts, Mamba-based architectures have yet to demonstrate consistent efficiency. As will be shown in the experimental section, these architectures exhibit unstable performance across medical imaging segmentation tasks. In some experiments, they achieve comparable performance with SOTA methods, while in others, they perform worse than baseline architectures, such as U-Net. Furthermore, the optimal placement of Mamba blocks remains uncertain. In this paper, we argue that Mamba blocks can be beneficial for fusion operations by enhancing the features and performing attention mechanisms within attention gates at a very low computational cost.

\begin{figure*}[h]
\begin{center}
\includegraphics[height=4.5in]{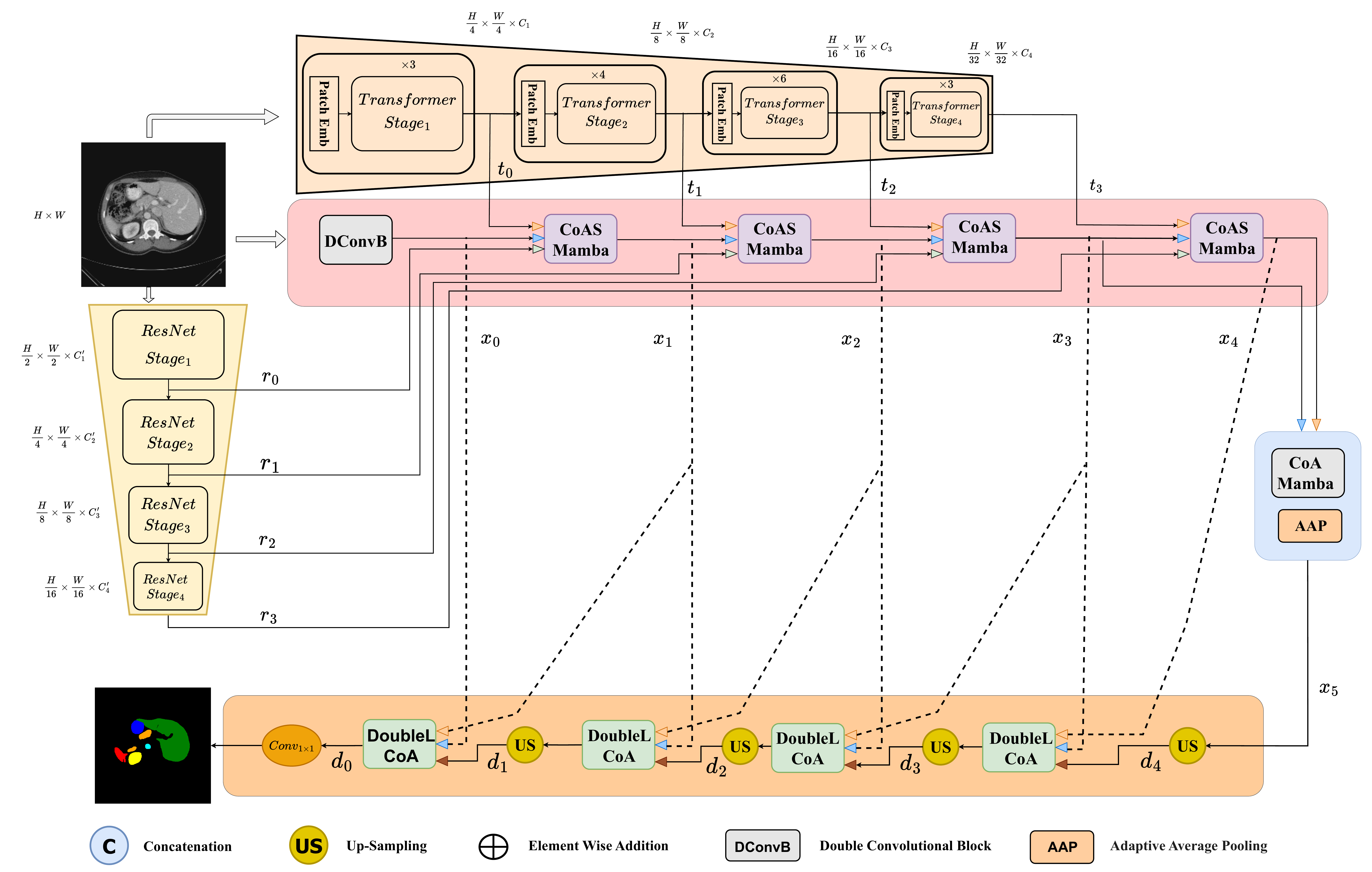} 
\caption{Our proposed MambaCAFU architecture. }
\label{fig:approach}
\end{center}
\end{figure*}
\begin{figure*}[h]
\begin{center}
    
\includegraphics[height=2in]{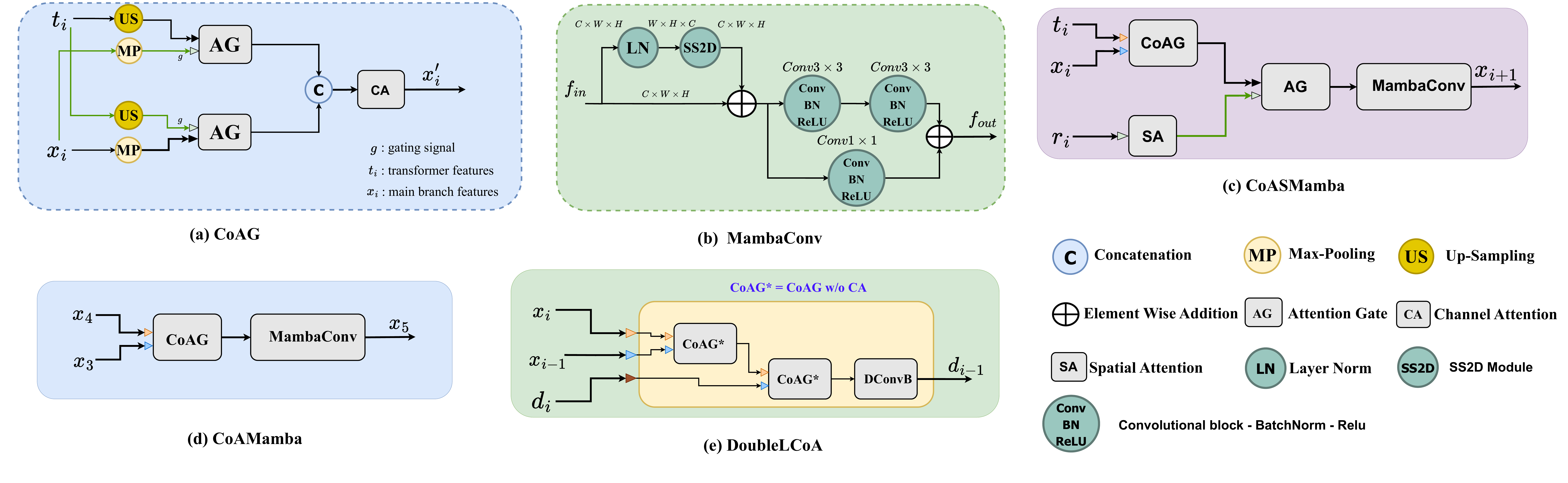} 
\caption{Detailed structure of our proposed blocks: (a) CoAG, (b) MambaConv, (c) CoASMamba, (d) CoAMamba and (e) DoubleLCoA. }
\label{fig:approach2}
\end{center}
\end{figure*}

\section{Proposed Approach}
\label{S:3}
As shown in Figure \ref{fig:approach}, our proposed MambaCAFU architecture integrates a hybrid CNN-Transformer-Mamba encoder with an attention-based CNN decoder. The encoder comprises three distinct branches: CNN, Transformer, and Mamba-based Attention Fusion. Each branch is fed by the input image. The CNN and Transformer branches extract pyramidal features across four stages, leveraging the complementary strengths of ResNet \cite{he2016deep} and PVT \cite{Wang_2021_ICCV, wang2022pvt} as backbones. CNNs excel in capturing local features, while Transformers are adept at modeling long-range dependencies. Building on prior efforts to fuse CNN and Transformer features at multiple scales \cite{zhang2021transfuse, yu2023unest, bougourzi2024rethinking}, we propose a novel fusion mechanism within the Mamba-based Attention Fusion (MAF) branch.

\subsection{Mamba-based Attention Fusion} As shown in Figure \ref{fig:approach}, the input image is fed into three paths: Transformer path, MAF path and CNN path. Both Transformer and CNN paths provide features flow at four stages, while MAF extracts and fuses the features within its six layers: a DConvB, four CoASMamba blocks, and a final CoAMamba block. 

In MAF, the input image is fed to the DConvB producing $x_0$, where DConvB includes two consecutive $3 \times 3$ convolutional blocks with a residual skip connection. The obtianed features from DConvB ($x_0$), along with the first-stage features from the Transformer ($t_0$) and CNN ($r_0$) branches, serve as input to the first CoASMamba block. Each CoASMamba block processes the features from the previous main branch features and the corresponding features from the Transformer and CNN branches, maintaining the main branch's flow while fusing them. These fused features, referred to as the main branch features ($x_1$, $x_2$, $x_3$, and $x_4$), are also passed to the decoder, as illustrated in Figure \ref{fig:approach}. Finally, the encoder features are further processed by our proposed CoAMamba layer that provides enhanced bottleneck features from two main branch features ($x_3$ and $x_4$).

\subsection{CoASMamba} 
In this section, we introduce the core module responsible for fusing Transformer features with MAF features. As shown in Figure \ref{fig:approach2}.c, CoAS-Mamba integrates a Co-Attention Gate (CoAG), a Mamba-based block (MambaConv), Spatial Attention (SpAtt), and an Attention Gate (AG) to enhance feature interaction and representation.

\noindent\textbf{Attention Gate:}

AG was proposed to be integrated into the decoder of the Att-Unet architecture, where it uses features from the previous encoder layer to identify salient regions in the encoder's skipped features \cite{oktay2018attention}. Recent research has shown that AG is also useful in constructing the encoder for medical image segmentation \cite{bougourzi2023pdatt, bougourzi2024rethinking}. The attention gate that has an input and gating signals $x$ and $g$, respectively, is defined by:

\begin{equation}
\label{eq:AG}
\hat{x} = AG(g,x) = x  \odot \alpha    
\end{equation}

\noindent where $\alpha$ is the attention coefficient matrix, which is defined by:

\begin{equation}
\alpha = \sigma_2 \left( Conv_{\psi} \left( \sigma_1 \left( Conv_x \, x + Conv_g \, \, g \right) \right) \right)
\end{equation}

\noindent where $\sigma_1$ and $\sigma_2$ are ReLu and Sigmoid activation functions, respectively. $Conv_x$, $Conv_g$ and $Conv_{\psi}$ are 1 by 1 convolutional blocks each followed by a batch normalization. These convolutional blocks has the weights $W_g \in \mathbb{R}^{F_g \times F_{int}}$, $W_x \in \mathbb{R}^{F_x \times F_{int}}$, and $W_{\psi} \in \mathbb{R}^{F_{int} \times 1}$, respectively.

\noindent\textbf{Co-Attention Gate (CoAG):} 
Our Proposed CoAG is designed to extract salient regions by using the Transformer and main branch features as interchangeable gating signals (Figure \ref{fig:approach2}.a). The resulting gated features are concatenated and further refined by a channel attention (CA) block, which emphasizes the most important channels from the gated features, as depicted in the following equations:

\begin{equation}
\label{eq:CoAtt}
\begin{split}
x_i'  &=  CoAG(x_i, t_i) \\
&= CA([AG_{i_1}(x_i, t_i), AG_{i_2}(t_i, x_i)]) 
\end{split}
\end{equation}

\noindent where: $CA$ denotes channel attention, and [] denotes concatenation. 

\noindent\textbf{MambaConv:} 
In order to encode long range dependencies with linear complexity, we integrated mamba module in each of the encoder's layer. As shown in Figure \ref{fig:approach2}.b, our proposed MambaConv consists of a mamba module followed by a residual block. In Mamba module, the input features $f_{in}$ are based through a Layer Norm (LN) followed by SS2D module, the obtained state-space features are then element-wisely summed with the input features $f_{in}$. To further extract deeper features, the residual block consists of two consecutive 3 by 3 convolutional blocks and summed with residual skipped features.  This module selects a prominent latent features from fused multi-branches and then CNN blocks extract local features from ranked outputs. In summary, MambaConv is defined by: 
\begin{equation}
\label{eq:MambaConv}
f_{out} = MambaConv(f_{in}) =ResB (f_{in} + SS2D(LN(f_{in}))) 
\end{equation}

\noindent where $SS2D$  is 2D-selective-scan module proposed in \cite{Liu2024VMambaVS}.  $LN$ and $ResB$  are layer norm and residual convolutional block, respectively.


\noindent\textbf{CoASMamba:} 
As shown in Figure \ref{fig:approach2}.c, CoASMamba takes three input features: transformer features $t_i$, main branch features $x_i$, and CNN features $r_i$. First, $t_i$ and $x_i$ are processed by the CoAG block to exchangeably localize prominent regions from each other, which is crucial for refining and combining features during the encoding phase. Meanwhile, $r_i$ is passed through spatial attention (SA) and then serves as a gating signal for the fused features from CoAG. This choice is based on the fact that both spatial attention and CNN features excel at capturing local patterns, so this double spatial awareness ensures enhanced spatial localization in each encoder layer. In order to extract long-range dependencies and further enhance the features with a low cost budget, MambaConv is applied on the output of the last AG. In summary, the CoASMamba is defined as follow:

\begin{equation}
\label{eq:CoASMamba}
\begin{split} 
\begin{aligned}
x_{i+1} &= CoASMamba(t_{i}, x_{i},r_i) \\ & = MambaConv( AG (SA(r_i), CoAG_i(t_{i}, x_{i})) 
\end{aligned}
\end{split}
\end{equation}

\noindent  where $SA$ is the spatial attention in \cite{10.1007/978-3-030-01234-2_1}.

\subsection{CoAMamba} \label{sssec:CoAMamba}
Since bottleneck features are the deepest representation of the encoder's features and will be mainly used to reconstruct the segmentation mask, we propose to exploit two features from the last two stages ($x_3$ and $x_4$). The aim of using two stages is to avoid missing any important information from the encoder. The features of the two stages are further processed using our proposed CoAMamba block. As shown in Figure \ref{fig:approach2}.d, CoAMamba is constructed using CoAG block followed by MambaConv block to encode the long-range dependencies features. In summary, CoAMamba is defined by:

\begin{equation}
\label{eq:CoAMamba}
\begin{split}
\begin{aligned}
    x_5 &= AAP(CoAMamba(x_3, x_4)) \\
        &=  AAP(MambaConv(CoAG(x_3,x_4)))
\end{aligned}
\end{split}
\end{equation}
where $AAP$ denotes adaptive average pooling with output dimension of 14 by 14. It should be noted that the CoAG used in CoAMamba is identical to the one described in Figure \ref{fig:approach2}.a, except that the MP on the input $x_3$ is removed.

\subsection{Co-Attention Based Decoder}
As depicted depicted in Figure \ref{fig:approach}, the decoder consists of four layer, each constructed by upsampling and DoubleLCoA block. The DoubleLCoA consists of modified co-attention blocks (CoAG*) and double convolutional block (DConvB).

\noindent\textbf{CoAG*:} the modified CoAG is identical to CoAG with removing CA block, as it has been found that CA and SA are effective only during the encoding phase and do not have significant during the decoding as the features are already enhanced during the encoding.

\noindent\textbf{DoubleLCoA:} 
As shown in Figure \ref{fig:approach2}.e, DoubleLCoA has two skip connection features and the bottleneck features for the first decoder layer or the previous decoder layer features as input. First, the two skipped features are passed through CoAG*, the obtained features are then passed through another CoAG* along side the decoder features. This aims to ensure passing complementary features from two encoder layers to the corresponding decoder. The first CoAG* block highlights the prominent features from encoder and the second CoAG* block interchangeability highlights the prominent features from both the encoder and decoder features. The obtained obtained features are further passed through double convolutional block (DConvB) to refine the obtained features. To obtain the segmentation mask, 1 by 1 convolutional block is used to match the features of the last decoder $e_0$ to match the number of classes.

\subsection{MambaCAFU Variants}
In evaluating our proposed MambaCAFU approach, we adopted two versions, which are  MambaCAFU-V\textsubscript{0} and MambaCAFU-V\textsubscript{1}. In which, the CNN branch is fixed to Resnet-18 and two versions of backbones are considered for Transformer branch, which are: PVTv2-B0 and PVTv2-B2-linear, respectively.

Table \ref{tab:demins} provides an overview of the MambaCAFU-V\textsubscript{1} architecture, detailing the input and output dimensions for each block.

\begin{table*}
\caption{Summary of our proposed MambaCAFU-V\textsubscript{1} architecture: layer composition with input and output dimensions.}
\label{tab:demins}
\centering
\begin{tabular}{l c c c}
\toprule
\textbf{Block} & \textbf{Layer} & \textbf{Input} & \textbf{Output} \\ 
\midrule
\multirow{9}{0.8in}{\textbf{Encoder}} 
& $DConvB$ & $I=[224 \times 224 \times 3]$ & $x_0=[224 \times 224 \times 32]$ \\ \cmidrule(l){2-4}
& $Transformer$ & $I=[224 \times 224 \times 3]$ & \makecell{$t_0=[56 \times 56 \times 64]$ \\ $t_1=[28 \times 28 \times 128]$ \\ $t_2=[14 \times 14 \times 320]$ \\ $t_3=[7 \times 7 \times 512]$ } \\ \cmidrule(l){2-4}
& $Resnet$ & $I=[224 \times 224 \times 3]$ & \makecell{$r_0=[112 \times 112 \times 64]$ \\ $r_1=[56 \times 56 \times 64]$ \\ $r_2=[28 \times 28 \times 128]$ \\ $r_3=[14 \times 14 \times 256]$} \\ \cmidrule(l){2-4}
& $CoASMamba_1$ & \makecell{$t_0=[56 \times 56 \times 64]$ \\ $x_0=[112 \times 112 \times 32]$ \\ $r_0=[112 \times 112 \times 64]$} & $x_1=[112 \times 112 \times 64]$ \\ \cmidrule(l){2-4}
& $CoASMamba_2$ & \makecell{$t_1=[28 \times 28 \times 128]$ \\ $x_1=[112 \times 112 \times 64]$ \\ $r_1=[56 \times 56 \times 64]$} & $x_2=[56 \times 56 \times 128]$ \\ \cmidrule(l){2-4}
& $CoASMamba_3$ & \makecell{$t_2=[14 \times 14 \times 320]$ \\ $x_2=[56 \times 56 \times 128]$ \\ $r_2=[28 \times 28 \times 128]$} & $x_3=[28 \times 28 \times 256]$ \\ \cmidrule(l){2-4}
& $CoASMamba_4$ & \makecell{$t_3=[7 \times 7 \times 512]$ \\ $x_3=[28 \times 28 \times 256]$ \\ $r_3=[14 \times 14 \times 256]$} & $x_4=[14 \times 14 \times 512]$ \\ \cmidrule(l){2-4}
& $CoAMamba_5$ & \makecell{$x_3=[28 \times 28 \times 256]$ \\ $x_4=[14 \times 14 \times 512]$} & $[28 \times 28 \times 512]$ \\ \cmidrule(l){2-4}
& $AAP$ & $[28 \times 28 \times 512]$ & $x_5=[14 \times 14 \times 512]$ \\ 
\midrule

\multirow{8}{0.8in}{\textbf{Decoder}} 
& $Up - Sampling_1$ & $x_5=[14 \times 14 \times 512]$ & $d_4=[28 \times 28 \times 512]$ \\ \cmidrule(l){2-4}
& $DoubleLCoA_1$ & \makecell{$x_3=[28 \times 28 \times 256]$ \\ $x_4=[14 \times 14 \times 512]$ \\ $d_4=[28 \times 28 \times 512]$} & $[28 \times 28 \times 256]$ \\ \cmidrule(l){2-4}
& $Up - Sampling_2$ & $[28 \times 28 \times 256]$ & $d_3=[56 \times 56 \times 256]$ \\ \cmidrule(l){2-4}
& $DoubleLCoA_2$ & \makecell{$x_2=[56 \times 56 \times 128]$ \\ $x_3=[28 \times 28 \times 256]$ \\ $d_3=[56 \times 56 \times 256]$} & $[56 \times 56 \times 128]$ \\ \cmidrule(l){2-4}
& $Up - Sampling_3$ & $[56 \times 56 \times 128]$ & $d_2=[112 \times 112 \times 128]$ \\ \cmidrule(l){2-4}
& $DoubleLCoA_3$ & \makecell{$x_1=[112 \times 112 \times 64]$ \\ $x_2=[56 \times 56 \times 128]$ \\ $d_2=[112 \times 112 \times 128]$} & $[112 \times 112 \times 64]$ \\ \cmidrule(l){2-4}
& $Up - Sampling_4$ & $[112 \times 112 \times 64]$ & $d_1=[224 \times 224 \times 64]$ \\ \cmidrule(l){2-4}
& $DoubleLCoA_4$ & \makecell{$x_0=[224 \times 224 \times 32]$ \\ $x_1=[112 \times 112 \times 64]$ \\ $d_1=[224 \times 224 \times 64]$} & $d_0=[224 \times 224 \times 32]$ \\ 
\bottomrule

\end{tabular}
\end{table*}



\section{Datasets}
\label{S:4}
We evaluate our approach on six widely used benchmark datasets covering diverse imaging modalities and segmentation tasks: Synapse \cite{landman2015miccai}, BTCV \cite{landman2015miccai}, ACDC \cite{bernard2018deep}, ISIC 2017 \cite{codella2018skin}, GlaS \cite{sirinukunwattana2017gland}, and MoNuSeg \cite{kumar2017dataset}.  

\subsection{Synapse Multi-Organ Segmentation}
The Synapse dataset consists of 30 abdominal CT scans from the MICCAI 2015 Multi-Atlas Abdomen Labeling Challenge, comprising 3,779 axial contrast-enhanced clinical CT images. Each volume contains 85–198 slices of size $512 \times 512$ pixels, with voxel resolutions ranging from ([0.54–0.54] $\times$ [0.98–0.98] $\times$ [2.5–5.0]) mm$^3$. Following the protocol in \cite{chen2021transunet}, we adopt the same training and testing splits, with 18 cases (2,212 slices) for training and 12 cases for validation. For evaluation, we report the average Dice Similarity Coefficient (DSC) and the 95\% Hausdorff Distance (HD95) across eight abdominal organs: aorta, gallbladder, spleen, left kidney, right kidney, liver, pancreas, and stomach as in  \cite{chen2021transunet}.  

\subsection{BTCV Multi-Organ Segmentation}  
The BTCV dataset extends the MICCAI 2015 Multi-Atlas Abdomen Labeling Challenge to 13 abdominal organs, namely: spleen, right kidney, left kidney, gallbladder, esophagus, liver, stomach, aorta, inferior vena cava (IVC), portal and splenic veins (P\&S vein), pancreas, right adrenal gland, and left adrenal gland. Following \cite{chen2021transunet}, we adopt the same training and testing splits with all 13 classes. Dataset preparation is consistent with \cite{rahman2024emcad}.\footnote{\url{https://github.com/SLDGroup/EMCAD}}

\subsection{ACDC Cardiac MRI Segmentation}  
The ACDC dataset collects cardiac MRI scans acquired using different scanners. Cine MR images were obtained under breath-hold conditions, covering the heart from the base to the apex of the left ventricle. Slice thickness ranges between 5–8 mm, with in-plane spatial resolution between 0.83–1.75 mm\textsuperscript{2}/pixel.  
Each scan is manually annotated with ground truth segmentations for the left ventricle (LV), right ventricle (RV), and myocardium (MYO). Following MT-UNet \cite{wang2022mixed}, the dataset is split into training, validation, and test sets. We report the average DSC across the three classes (LV, RV, MYO). Dataset preparation follows \cite{rahman2024emcad}.\footnote{\url{https://github.com/SLDGroup/EMCAD}}  

\subsection{Skin Lesion Segmentation (ISIC 2017)}  
The ISIC 2017 dataset \cite{codella2018skin} contains training, validation, and test images for skin lesion segmentation. Following prior works, we report accuracy (ACC), intersection-over-union (IoU), and Dice coefficient for binary segmentation. The dataset is publicly available.\footnote{\url{https://challenge.isic-archive.com/data/\#2017}}  

\subsection{Gland and MoNuSeg Segmentation}  
The Gland Segmentation dataset (GlaS) \cite{sirinukunwattana2017gland} contains 85 training and 80 test images, while the MoNuSeg dataset \cite{kumar2017dataset} includes 30 training and 14 test images. Following UCTransNet \cite{UCTransNet}, we report the mean $\pm$ standard deviation (std) of DSC and IoU over three runs of 5-fold cross-validation.  
The GlaS dataset is available online,\footnote{\url{https://drive.google.com/file/d/1t2MDLkj5DYyGYBGeT7Q7u8F6rscqNzCF/view?usp=drive_web}} and the MoNuSeg dataset can be accessed at the UCTransNet repository.\footnote{\url{https://github.com/McGregorWwww/UCTransNet/tree/main/datasets/MoNuSeg}}


\section{Evaluation Metrics and Training Settings}
\label{S:5}
\subsection{Metrics}
The performance of the methods was evaluated using the following evaluation metrics, which are commonly used in state-of-the-art works and following the same evaluation metric used by SOTA methods in each dataset.

\noindent\textbf{Accuracy}: It is calculated as the ratio of correctly classified pixels to the total number of pixels.

\begin{equation}
    \text{Accuracy} = \frac{\sum_{i=1}^{C}x_{ii}}{\sum_{i=1}^{C}\sum_{j=1}^{C}x_{ij}}
\end{equation}

\noindent where $C$ is the number of classes, $x_{ii}$ is the number of pixels of class $i$ correctly predicted, and $x_{ij}$ is the number of pixels of class $i$ predicted as class $j$.

\noindent\textbf{Average Dice Score (DSC)}: This metric is widely used in medical image segmentation. It is defined as twice the overlap area between the predicted segmentation map and the ground truth annotations, divided by the total number of pixels in both maps.

\begin{equation}
    DSC(A,B) = \frac{2\left| A \cap B \right|}{\left| A \right| + \left| B \right|}
\end{equation}

\noindent where $A$ and $B$ represent the ground truth and predicted segmentation maps, respectively. The DSC ranges from 0 to 1, with higher values indicating better performance. The mean DSC is computed as the average DSC across all target classes. In the case of a binary segmentation task, the DSC score is equivalent to the F1 score, which is computed as:

\begin{equation}
    F1 = \frac{2TP}{2TP + FP + FN}
\end{equation}

\noindent where $TP$, $FP$, and $FN$ denote the true positive, false positive, and false negative counts, respectively.

\noindent\textbf{Hausdorff Distance (HD)}: The maximum Hausdorff distance measures the largest distance from any point in the predicted segmentation map $B$ to the closest point in the ground truth segmentation $A$. Formally, it is defined as:

\begin{equation}
    \max\left\{ d_{AB}, d_{BA} \right\} = \max \left\{ \max\limits_{x \in A} \min\limits_{y \in B} d(x,y), \max\limits_{y \in B} \min\limits_{x \in A} d(x,y) \right\}
\end{equation}

\noindent where $HD_{95}$ refers to the 95th percentile of the distances between boundary points in $A$ and $B$.

\noindent\textbf{Intersection over Union (IoU)}: IoU is defined as the area of intersection between the predicted segmentation map and the ground truth, divided by the area of their union:

\begin{equation}
    IoU(A,B) = \frac{\left| A \cap B \right|}{\left| A \cup B \right|}
\end{equation}

\noindent where $A$ and $B$ denote the ground truth and predicted segmentation maps, respectively. IoU ranges from 0 to 1, with higher values indicating better performance. The mean IoU is computed as the average IoU across all target classes.

\subsection{Training Setup}
\subsubsection{abdominal Multi-organs Segmentation:} 

For training our model using the Synapse dataset, we set the initial learning rate to 0.0025 and batch size to 18, employing the AdamW optimizer along with a cosine annealing learning rate scheduler with warm restarts (with the restart period $T$ set to 2). The weighted loss function is defined as $\mathcal{L} = 0.8 \times \mathcal{L}{dice} + 0.2 \times \mathcal{L}{bce}$. Since the Synapse dataset lacks validation data, we trained our architecture for 100 epochs and evaluated the final model on the testing data. Similarly, for training on the BTCV dataset, we used the same optimizer process as in the Synapse training with setting the initial learning rate to 0.003, and employed a weighted loss function defined as $\mathcal{L} = 0.6 \times \mathcal{L}{dice} + 0.4 \times \mathcal{L}{bce}$.

For the experiments of PVT-EMCAD \cite{rahman2024emcad} and MERIT \cite{rahman2024multi} architectures, we reevaluate these methods using their official implementation. The original training settings of these works are preserved, only following the same evaluation protocol as in \cite{chen2021transunet}. In this experiments, we selected the optimal number of epochs considering both Synapse and BTCV datasets. Thus, the number of epochs are set to 100 and 120 for MERIT and EMCAD, respectively.  

For the mamba based segmentation architectures (Swin-UMamba, UMamba \cite{U-Mamba}, and Mamba-UNet \cite{wang2024mamba}), we evaluated these architectures using the same training setup as our proposed  MambaCAFU including batch size, epoch, initial learning rate and the loss function for fair comparison.

\subsubsection{Automated Cardiac Diagnosis challenge} 

ACDC supports three parts of training, validation and testing. Thus, the validation data is used to select the training hyperparameters with considring the best DSC. The batch size, initial learning rate, and the the number of epochs are set to 12, 0.01, and 400, respectively.  The loss function is set to $\mathcal{L} = 0.6 \times \mathcal{L}_{dice} + 0.4 \times \mathcal{L}_{bce}$. 

On this dataset, we evaluate two Mamba-based architectures: Mamba-Unet and Swin-Umamba using the same evaluation settings as the ones used to train our proposed approach.

\subsubsection{Skin lesion segmentation} 
In the training of this task, we used a batch size and initial learning rate of 6 and 0.01, respectively. The used loss function is $\mathcal{L} = 0.6 \times \mathcal{L}_{dice} + 0.4 \times \mathcal{L}_{bce}$. Also, we evaluated two Mamba-based architectures (Mamba-Unet and Swin-Umamba) using the same evaluation settings as used to train our proposed approach.

\subsubsection{Gland and MoNuSeg segmentation} 
During the training process, the batch size, initial learning rate, and the number of epochs are set to 16, 0.1 and 100, respectively. The used optimizer is Adam. We fuse the dice  and cross-entropy loss functions with equal weight of 0.5: $\mathcal{L} = 0.5 \times \mathcal{L}_{dice} + 0.5 \times \mathcal{L}_{bce}$. 
For the evaluated Mamba-based architectures, the initial learning rate is set to 1e-4, which showed better performance for these methods.



\begin{table*}
\caption{Comparison of Abdominal Multi-Organ Segmentation on the Synapse Dataset. DSC and HD95 represent the average Dice Score and 95\% Hausdorff Distance across eight classes, respectively. Columns 4 to 11 present the Dice Score for each class, while the last two columns indicate model complexity in terms of the number of parameters (millions) and the number of FLOPs (GMac), respectively. * denotes models that we reevaluated using their original evaluation settings while adhering to the evaluation protocol and data split from \cite{chen2021transunet} for a fair comparison. ** marks a recent Mamba-based model that we evaluated on the Synapse Dataset due to the absence of prior results for this approach on this dataset. Bold values indicate the best result, while underlined values denote the second-best result.}
\begin{center}
\label{tab:synapse}
\centering
\resizebox{2\columnwidth}{!}{
\begin{tabular}{|l||c|c||c|c|c|c|c|c|c|c||c|c|}

\hline
 {\multirow{2}{*} \textbf{Architecture}}    & \multicolumn{2}{|c|}{\textbf{Average}}& \multirow{2}{*} \textbf{Aorta} &\multirow{2}{*} \textbf{Gallbladder} &\multirow{2}{*} \textbf{Kidney (L) } &\multirow{2}{*} \textbf{Kidney (R)}& \multirow{2}{*} \textbf{Liver}& \multirow{2}{*} \textbf{Pancreas}& \multirow{2}{*} \textbf{Spleen}& \multirow{2}{*} \textbf{Stomach} & \multicolumn{2}{|c|}{\textbf{Complexity}} \\
       
\cline{2-3} \cline{12-13}
& \textbf{DSC$\uparrow$}& \textbf{HD95$\downarrow$} & &   & &  & &   & &  & \textbf{\#Params}& \textbf{\#FLOPs} \\
\hline\hline 

Unet \cite{10.1007/978-3-319-24574-4_28}&  74.68 &36.87& 84.18& 62.84 &79.19& 71.29& 93.35 &48.23 &84.41 &73.92 & 7.85 & 10.73 \\\hline

Att-Unet \cite{oktay2018attention}&75.57& 36.97& 55.92& 63.91& 79.20& 72.71 &93.56& 49.37& 87.19 &74.95 & 7.98 & 11.05 \\\hline \hline

TransUnet \cite{chen2021transunet}& 77.48& 31.69 &87.23 &63.13 &81.87& 77.02 &94.08& 55.86 &85.08 &75.62 & 105.28 & 24.66  \\\hline

MTUnet \cite{wang2022mixed}& 78.59 & 26.59 &  87.92&   64.99& 81.47& 77.29 & 93.06& 59.46  &  87.75 & 76.81 & 79.07 & 44.73 \\\hline

UCTransNet \cite{wang2022uctransnet}& 78.23 & 26.75 & \bf{-} &  \bf{-} & \bf{-}& \bf{-} & \bf{-}&  \bf{-} &  \bf{-} & \bf{-} & 66.43 &32.94  \\\hline

TransClaw U-Net \cite{wang2022uctransnet}& 78.09 & 26.38 &85.87 & 61.38  & 84.83& 79.36 & 94.28&  57.65 &  87.74 &  73.55 & \bf{-} &  \bf{-} \\\hline
          
ST-Unet \cite{zhang2023st} &78.86 &20.37 &85.68& 69.05& 85.81&  73.04&  95.13&60.23&   89.15&  72.78 & \bf{-} &  \bf{-}
\\\hline

Swin-Unet \cite{liu_swin_2021} &77.58 &27.32 & 81.76 &65.95 &82.32 &79.22 &93.73& 53.81 &88.04 &75.79 & 41.38 &15.12  \\\hline

VM-UNet \cite{ruan2024vm}& 81.08& 19.21& 86.40 &69.41& 86.16 &82.76& 94.17 &58.80 &89.51 &81.40 & \bf{-} &  \bf{-} \\\hline

TransCeption \cite{azad2023enhancing}& 82.24& 20.89 &87.60 &71.82& 86.23 &80.29 &95.01 &65.27 &91.68 &80.02 & \bf{-} &  \bf{-}
\\\hline

PVT-EMCAD-B2* \cite{rahman2024emcad}& 82.66&  19.35 &85.61 & 67.28 &86.31 & 83.94  &95.22 & 67.28 &91.12 & \underline{84.51} & 26.77 & 4.44  \\\hline

Parallel MERIT* \cite{rahman2024multi}& 82.43&  16.20 &87.37 & 69.03 &85.35 & 82.48  &94.97 & 67.05 &90.18 & 83.02 & 147.85  & 34.19 \\\hline

Cascaded MERIT* \cite{rahman2024multi}& 83.45&  \bf{13.24} &87.87 & \bf{73.93} &84.71 & 83.40  &94.85 & 67.82 &90.76 & 84.27 & 147.86  & 33.31 \\\hline

PAG-TransYnet \cite{bougourzi2024rethinking}& 83.43&  \underline{15.82} &89.67 & 68.89 &\underline{86.74} & \bf{84.88}  &\underline{95.87} & \underline{68.75} &\underline{92.01} & 80.66  & 144.22 &33.65   \\\hline

\hline

UMamba\_Bot\_2D** \cite{U-Mamba}& 74.74&  29.26 &86.37 & 59.45 &81.50 & 72.29  &93.82 & 51.12 &88.20 & 65.14 & 21.33 & 11.26  \\\hline

UMamba\_Enc\_2D** \cite{U-Mamba}& 76.30&  27.23 &87.08 & 59.36 &78.60 & 70.14  &93.63 & 58.14 &88.48 & 74.98 & 21.62 & 9.95   \\\hline

Mamba-Unet** \cite{wang2024mamba}& 76.21&  22.64 &85.40 & 65.88 &84.51 & 76.65  &93.46 & 49.50 &85.68 & 68.61 & 19.12 & 4.53   \\\hline

Swin-UMamba** \cite{linguraru_swin-umamba_2024}& 80.34&  21.51 &87.46 & 66.45 &84.39 & 78.49  &95.17 & 63.35 &90.08 & 77.30 & 59.88 & 31.35   \\\hline

SliceMamba \cite{fan2025slicemamba}& 81.95& 16.04  &87.78  &68.77 &\bf{88.30} &84.26  &95.25 &64.49  &86.91 &79.82  & - &  -  \\\hline\hline

MambaCAFU-V\textsubscript{0} & 82.81&  18.20 &\bf{90.22} & \underline{72.90} & 86.39 & 83.79  & 95.33 & 64.33 & 89.93 & 79.62 & 42.40 & 33.59  \\\hline

MambaCAFU-V\textsubscript{1}  & \bf{84.87}&  17.15 &\underline{89.75} & 71.69 &\underline{87.84} & \underline{84.80}  &\bf{96.17} & \bf{70.93} &\bf{92.50} & \bf{85.32} &  66.71 &40.31  \\\hline

\end{tabular}}
\end{center}
\end{table*}

\begin{table*}
\caption{Comparison of Abdominal Multi-Organ Segmentation on the BTCV Dataset. DSC and HD95 are the average of dice score and 95\% Hausdorff distance of the 13 classes, respectively. Columns four to the last show the Dice-score for each class.  Bold indicates the best result, and underline indicates the second-best result. }
\begin{center}
\label{tab:BTCV}
\centering
\resizebox{2\columnwidth}{!}{
\begin{tabular}{|l||c|c||c|c|c|c|c|c|c|c|c|c|c|c|c|}
\hline
\multirow{2}{*}{Architecture} & \multicolumn{2}{c|}{\textbf{Average}}       & \multirow{2}{*}{Spleen} & \multirow{2}{*}{Kidney (R)} & \multirow{2}{*}{Kidney (L)} & \multirow{2}{*}{Gallbladder} & \multirow{2}{*}{Esophagus} & \multirow{2}{*}{Liver} & \multirow{2}{*}{Stomach} & \multirow{2}{*}{Aorta} & \multirow{2}{*}{IVA} & \multirow{2}{*}{P\&S Vein} & \multirow{2}{*}{Pancreas} & \multirow{2}{*}{AG (R)} & \multirow{2}{*}{AG (L)} \\ \cline{2-3}
 & \multicolumn{1}{l|}{\textbf{DSC$\uparrow$}}   & \textbf{HD95$\downarrow$}    &             &        &      &               &       &      &       &                &    &                       &       &    &      \\ \hline
U-Net \cite{10.1007/978-3-319-24574-4_28}& 72.86 & 20.88 & 84.22     & 71.95 & 80.23 & 62.05 & \underline{73.00} & 92.81 & 70.74 & \bf{88.64} & 77.96 & 62.53 & 59.95   & 59.78  & \underline{63.27}  \\ \hline
UCTransNet \cite{UCTransNet}  & 72.45 & 24.38 & 86.09  & 70.38   & 80.62  & 63.47  & \bf{74.27}  & 93.41  & 74.17  & 86.52   & 76.81  & 62.27 & 55.00  & \underline{61.27} & 57.60    \\ \hline
PVT-EMCAD-B2 \cite{rahman2024emcad}  & 72.41 & 14.52 & 88.95    & 79.27 & 83.92  & 66.12  & 71.44   & \underline{95.50}  & 81.49   & 86.06   & 75.67  & 64.71           & 64.09  & 45.95   & 38.12   \\ \hline 
Cascaded MERIT \cite{rahman2024multi}  & 72.74 & 16.10 & 89.45 & 76.38  & 80.36 & \underline{68.03} & 70.03  & 94.37  & 79.60   & 85.12 & 71.88  & 65.79  & 61.72  & 53.01    & 49.88    \\ \hline
Parallel MERIT \cite{rahman2024multi}  & 74.78 & \underline{13.53} & \underline{90.70} & \underline{83.00} & \underline{86.17}  & \bf{68.30}  & 68.42   & 94.60  & \bf{82.70}   & 85.48   & 74.83  & \bf{68.57}  & \bf{67.04}  & 50.91   & 51.40    \\ \hline
\hline
UMamba\_Bot\_2D \cite{U-Mamba}  & 67.30 & 25.23 & 84.40     & 71.05  & 79.22  & 47.30   & 63.82  & 93.43 & 69.07   & 86.07  & 69.32    & 53.54  & 56.72   & 51.19    & 49.82    \\ \hline
UMamba\_Enc\_2D \cite{U-Mamba}  & 65.97 & 14.97 & 87.37     & 69.82  & 78.25  & 59.55   & 67.27  & 94.05   & 72.81   & 85.31  & 72.37  & 58.96  & 59.48   & 52.39    & 00.00    \\ \hline 
Mamba-Unet \cite{wang2024mamba}  & 69.46& 14.47 & 90.42     & 78.06 & 83.67  & 65.02  & 65.68   & 94.30  & 71.48   & 82.01   & 69.10  & 56.99  & 51.16  & 55.39   & 39.70       \\ \hline
Swin-UMamba \cite{linguraru_swin-umamba_2024}  & 74.56 & 16.17 & 88.97  & 81.04 & 85.09  & 67.28  & 71.91   & 95.30  & 77.24   & 86.85   & 78.75  & 64.22           & 63.55  & 56.72   & 52.40    
\\ \hline \hline
MambaCAFU-V\textsubscript{0}   & \underline{75.84} & \bf{12.50} & \bf{91.06}  & \bf{84.29} & \bf{87.07}  & 61.99 & 70.67 & 95.38  & 79.28  & \underline{88.25}  & \bf{79.58}   & 66.46    & 65.61  & 56.51   & 59.82 \\ \hline
MambaCAFU-V\textsubscript{1}   & \bf{76.86} & 16.63 & 89.92  & 80.27& 85.16  & 64.94        & 72.81 & \bf{95.77}  & \underline{81.59}  & 87.92  & \underline{78.79}   & \underline{68.47}    & \underline{65.75}  & \bf{63.98}   & \bf{63.79} \\ \hline
\end{tabular}}
\end{center}
\end{table*}


\begin{table}
\caption{Comparison on Automatic Cardiac Diagnosis Challenge (ACDC) dataset. Bold indicates the best result, and underline indicates the second-best result.   }
\label{tab:ACDC}
\centering
\resizebox{1\columnwidth}{!}{\begin{tabular}{|l|l|c|c|c|c|}
\hline
Ex & Architecture   & \textbf{Avg DSC} & RV    & Myo   & LV    \\ \hline
1  & TransUNet \cite{chen2021transunet}     & 89.71   & 86.67 & 87.27 & 95.18 \\ \hline
2  & SwinUNet \cite{cao2022swin}      & 88.07   & 85.77 & 84.42 & 94.03 \\ \hline
3  & MT-Unet \cite{wang2022mixed}       & 90.43   & 86.64 & 89.04 & 95.62 \\ \hline
4  & MISSFormer \cite{huang2022missformer}     & 90.86   & 89.55 & 88.04 & 94.99 \\ \hline
5  & PVT-CASCADE \cite{rahman2023medical}   & 91.46   & 89.97 & 88.90 & 95.50 \\ \hline
6  & TransCASCADE \cite{rahman2023medical}   & 91.63   & 90.25 & 89.14 & 95.50 \\ \hline
7  & Cascaded MERIT \cite{rahman2024multi} & 91.85   & 90.23 & 89.53 & 95.80 \\ \hline
8  & PVT-EMCAD-B0 \cite{rahman2024emcad}   & 91.34   & 89.37 & 88.99 & 95.65 \\ \hline
9  & PVT-EMCAD-B2 \cite{rahman2024emcad}   & \underline{92.12}   & \underline{90.65} & \underline{89.68} & \underline{96.02} \\ \hline
10  & Mamba-Unet \cite{wang2024mamba}   & 83.75   & 81.52 & 78.49 & 91.25 \\ \hline 
11  & Swin-UMamba \cite{linguraru_swin-umamba_2024}   & 91.18   & 89.34 & 88.94 & 95.26 \\ \hline \hline 
12  & MambaCAFU-V\textsubscript{0}   & 91.34   & 88.86 & 89.22 & 95.93 \\ \hline
13 & MambaCAFU-V\textsubscript{1}       & \textbf{92.37}   & \textbf{90.68} & \textbf{90.32} & \textbf{96.10} \\ \hline
\end{tabular}}
\end{table}



\begin{table}
\caption{Comparison on Skin lesion ISIC17 dataset. Comparison results from 1-9 are obtained from \cite{WU2022102327} and results from 10-14 are obtained from \cite{rahman2024emcad}. Bold indicates the best result, and underline indicates the second-best result. }
\label{tab:ISIC2017}
\centering
\begin{tabular}{|l|l||c|c|c|}
\hline
Ex & Architecture &  \textbf{ACC}   & \textbf{IoU}   & \textbf{Dice}  \\ \hline
1  & U-Net \cite{10.1007/978-3-319-24574-4_28}        & 91.64 & 72.34 & 81.59 \\ \hline
2  & AttU-Net \cite{SCHLEMPER2019197}     & 91.45 & 71.73 & 80.82 \\ \hline
3  & CPFNet \cite{9049412}       & 92.15 & 75.46 & 84.03 \\ \hline
4  & ERU \cite{9303084}          & 91.98 & 75.18 & 84.13 \\ \hline
5  & SESV \cite{9201384}         & 92.23 & 5.31  & 83.92 \\ \hline
6  & MB-DCNN \cite{8990108}      & 93.11 & 76.03 & 84.27 \\ \hline
7  & DAGAN \cite{LEI2020101716}       & 93.04 & 75.94 & 84.25 \\ \hline
8  & TransUNet \cite{10484516}  & 92.07 &  - & 81.23 \\ \hline
9  & FAT-Net \cite{WU2022102327}      & 93.26 & 76.53 & 85.00 \\ \hline \hline

10  & Swin-UNet \cite{cao2022swin}   & -   & - & 83.97\\ \hline
11  & TransFuse \cite{zhang2021transfuse}   & -   & - & 84.89 \\ \hline
12  & PVT-CASCADE \cite{rahman2023medical}  & -   & - & 85.50 \\ \hline
13  & PVT-EMCAD-B0 \cite{rahman2024emcad}    & -   & - &85.67  \\ \hline
14  & PVT-EMCAD-B2 \cite{rahman2024emcad}    & -   & - & \underline{85.95} \\ \hline 
15  & Mamba-Unet \cite{wang2024mamba}   & 93.46   & 76.24 & 84.56\\ \hline
16  & Swin-UMamba \cite{linguraru_swin-umamba_2024}   & 93.95   & 77.30 & 85.47 \\ \hline \hline

17  & MambaCAFU-V\textsubscript{0}     & 93.66 & 77.08 & 85.43 \\ \hline
18 &MambaCAFU-V\textsubscript{1}     & \textbf{94.07} & \textbf{78.27} & \textbf{86.26} \\ \hline
\end{tabular}
\end{table}

\begin{table*}
\caption{Comparison on GlaS and  MoNuSeg Segmentation datasets. The comparison results 1-7 are obtained from \cite{wang2022uctransnet}. Bold indicates the best result, and underline indicates the second-best result.  }
\label{tab:GlaSMoNuSeg}
\centering
\begin{tabular}{|c|l|c|c||c|c|}

\hline
 Ex &{\multirow{2}{*}    \textbf{Architecture}}    & \multicolumn{2}{|c|}{\textbf{GlaS}}& \multicolumn{2}{|c|}{\textbf{MoNuSeg}} \\


&  &\textbf{DSC} &   \textbf{IoU} &\textbf{DSC} &   \textbf{IoU}  \\
\hline

1& U-Net \cite{10.1007/978-3-319-24574-4_28}& 85.45$\pm$1.3  &  74.78$\pm$1.7  & 76.45$\pm$2.6 & 62.86$\pm$3.0 \\\hline 

2& Unet++ \cite{zhou_unet_2018}& 87.56$\pm$1.2  &  79.13$\pm$1.7  & 77.01$\pm$2.1 & 63.04$\pm$2.5\\\hline 

3& AttUNet \cite{oktay2018attention}& 88.80$\pm$1.1  &  80.69$\pm$1.7  & 76.67$\pm$1.1 & 63.47$\pm$1.2 \\\hline 


4& TransUNet \cite{chen2021transunet}& 88.40$\pm$0.7  &  80.40$\pm$1.0  & 78.53$\pm$1.1 & 65.05$\pm$1.3 \\\hline

5& MedT \cite{10.1007/978-3-030-87193-2_4}& 85.92$\pm$2.9  &  75.47$\pm$3.5  & 77.46$\pm$2.4 & 63.37$\pm$3.1 \\\hline

6& Swin-Unet \cite{cao2022swin}& 89.58$\pm$0.6  &  82.06$\pm$ 0.7 & 77.69$\pm$0.9 & 63.77$\pm$ 1.2\\\hline

7& UCTransNet \cite{wang2022uctransnet}& 90.18$\pm$0.7  &  82.96$\pm$1.1  & 79.08$\pm$0.7 & 65.50$\pm$0.9 \\\hline


8& PAG-TransYnet \cite{bougourzi2024rethinking}& 94.20$\pm$0.55 & 89.29$\pm$0.91& 79.62$\pm$0.7& 66.31$\pm$0.6 \\\hline 

9& Mamba-Unet \cite{wang2024mamba}& 93.41$\pm$3.01 & 87.98$\pm$5.02& 76.54$\pm$1.00& 62.15$\pm$1.36 \\\hline 

10& Swin-UMamba \cite{linguraru_swin-umamba_2024}& 95.85$\pm$2.48 & 92.30$\pm$4.20& \underline{81.45$\pm$2.14}& \underline{68.87$\pm$3.05} \\\hline \hline

11& MambaCAFU-V\textsubscript{0}  & \underline{96.21$\pm$1.54} & \underline{92.82$\pm$2.70}& 80.27$\pm$0.8 & 67.19$\pm$1.1 \\\hline 

12& MambaCAFU-V\textsubscript{1}  & \bf{96.76$\pm$1.50} & \bf{93.82$\pm$2.80}& \bf{81.85$\pm$0.7}& \bf{69.40$\pm$1.0} \\\hline 
\end{tabular}
\end{table*}


\section{Experimental Results}
\label{S:6}
\subsection{Implementation settings}
We implemented our model with PyTorch using NVIDIA V100 GPU card with 32 GB memory. Standard data augmentation are used during the training including horizontal flipping, vertical flipping and random rotating a degree of -20 to 20.  We adopted weighted loss function as follow $\mathcal{L} = \alpha \mathcal{L}_{dice} + (1-\alpha) \mathcal{L}_{bce}$. We employed Adam optimizer, cosine learning rate schedule with warm restart.

\subsection{Results}
Tables \ref{tab:synapse}, \ref{tab:BTCV}, \ref{tab:ACDC}, \ref{tab:ISIC2017}, and \ref{tab:GlaSMoNuSeg} summarize the evaluation of our proposed approach compared with SOTA architectures across  Synapse  \cite{landman2015miccai}, BTCV \cite{landman2015miccai}, ACDC \cite{bernard2018deep},  ISIC17 \cite{codella2018skin}, and GlaS  \cite{sirinukunwattana2017gland}, and MoNuSeg \cite{kumar2017dataset}, respectively.

Table \ref{tab:synapse} presents the comparison of the segmentation performance of our method with CNN-based models, a pure Transformer architecture (Swin-Unet), hybrid Transformer-CNN models, and Mamba-based approaches on the Synapse dataset. Our MambaCAFU-V\textsubscript{1} achieves the highest average Dice score across the eight classes and the third-best HD95. Furthermore, Mamba-based architectures generally show moderate performance, being outperformed by multiple hybrid CNN-Transformer models. Our proposed  MambaCAFU-V\textsubscript{1} surpass the best Mamba-based comparison model (SliceMamba) by 2.92 in terms of DSC metric, 

In terms of individual class segmentation, our approach attains the best Dice score in four classes and the second-best in three others among a total of 8 classes. MambaCAFU-V\textsubscript{0} also demonstrates strong performance compared to CNN, Transformer, and Mamba-based models, remaining competitive with leading hybrid Transformer-CNN approaches. Notably, it achieves the best segmentation performance for the Aorta and the second-best for the Gallbladder. Regarding computational efficiency, our model maintains a moderate parameter count (42M–66M) and FLOPs (33G–40G).

\begin{table*}
\caption{Ablation study on Synapse and ISIC17 Datasets. The importance of the following elements is studied: ResNet branch (ResB), CoAG and MambaConv. }

\label{tab:abblocks}
\centering
\begin{tabular}{|l||ccc||c||c|}
\hline
{\multirow{2}{*}    \textbf{Architecture}}   &\multicolumn{3}{|c|}{\textbf{Ablation}}  & \textbf{Synapse}&\textbf{ISIC17}  \\

 &\textbf{ResB} &\textbf{CoAG}&\textbf{MambaCov} &\textbf{DSC} &  \textbf{DSC} \\
\hline




\bf{1. Baseline} &\xmark& \xmark & \xmark & 82.54 &   83.93 \\\hline

\bf{2. w/o MambaCov} &\cmark& \cmark & \xmark & 83.76  & 85.82\\\hline

%

\bf{3. w/o ResBranch} &\xmark& \cmark &  \cmark & 83.43   & 85.60\\\hline

\bf{4. w/o CoAG, CoAG*} &\cmark& \xmark &  \cmark &  83.97  & 86.30 \\\hline

\bf{5. MambaCAFU-V1} &\cmark& \cmark &  \cmark& 84.87   & 86.26\\\hline

\end{tabular}
\end{table*}


\begin{table*}
\caption{Ablation study on Synapse Dataset and ISIC17. The importance of the following elements is studied: CoASMamba, CoAMamba and DoubleLCoA. } 
\label{tab:ablayers}
\centering
\resizebox{1 \textwidth}{!}{
\begin{tabular}{|l||ccc||c||c|}
\hline

{\multirow{2}{*}    \textbf{Architecture}}   &\multicolumn{3}{|c|}{\textbf{Ablation}}  & \textbf{Synapse}&\textbf{ISIC17}  \\


&\textbf{CoASMamba} &\textbf{CoAMamba}&\textbf{DoubleLCoA} &\textbf{DSC} &\textbf{DSC} \\
\hline




\bf{1. Baseline} &\xmark& \xmark  & \xmark& 82.65 &   83.90 \\\hline

\bf{2. Add CoASMamba} &\cmark& \xmark  & \xmark& 83.25  & 85.99 \\\hline

\bf{3. Add CoASMamba and CoAMamba} &\cmark&   \cmark & \xmark& 83.65  & 86.08 \\\hline

\bf{4. MambaCAFU-V1} &\cmark& \cmark &  \cmark & 84.87   & 86.26 \\\hline


\end{tabular}}
\end{table*}


Table \ref{tab:BTCV} presents the results for abdominal multi-organ segmentation on the BTCV dataset. Given the absence of a unified evaluation protocol, we follow the dataset split from \cite{chen2021transunet}, segmenting thirteen abdominal organs to ensure a fair comparison. Our MambaCAFU-V\textsubscript{0} and MambaCAFU-V\textsubscript{1} achieve the second-best and best average DSC across the thirteen classes, respectively. In HD95, MambaCAFU-V\textsubscript{0} achieves the best performance.

Segmenting a larger number of abdominal organs (13 classes) presents greater challenges than segmenting the Synapse dataset (8 classes). Interestingly, the baseline U-Net \cite{10.1007/978-3-319-24574-4_28} performs comparably to several hybrid Transformer-CNN models and surpasses three SOTA Mamba-based architectures. This suggests that these methods are highly task-dependent, with varying performance across different datasets. Nevertheless, our approach consistently outperforms SOTA methods, improving over U-Net by 2.98\% and 4\% in DSC with MambaCAFU-V\textsubscript{0} and MambaCAFU-V\textsubscript{1}, respectively.

Table \ref{tab:ACDC} presents results for cardiac anatomical structure segmentation on the ACDC dataset. Our MambaCAFU-V\textsubscript{1} achieves the best average Dice score and the highest Dice score for all classes. Specifically, MambaCAFU-V\textsubscript{1} outperforms PVT-EMCAD-B2 and Cascaded-MERIT by 0.25\% and 0.52\% in average Dice score, respectively.

For binary segmentation tasks, Tables \ref{tab:ISIC2017} and \ref{tab:GlaSMoNuSeg} demonstrate that our approach surpasses SOTA methods in segmenting skin cancer, glands, and multi-organ nuclei across the ISIC17, GlaS, and MoNuSeg \cite{kumar2017dataset} datasets. Specifically, MambaCAFU-V\textsubscript{1} outperforms U-Net by 4.67\% and the best comparison method, PVT-EMCAD-B2, by 0.31 in DSC on the ISIC17 dataset. Additionally, MambaCAFU-V\textsubscript{0} and MambaCAFU-V\textsubscript{1} achieve the second-best and best performance, respectively, with MambaCAFU-V\textsubscript{1} outperforming the best SOTA method by 2.56\% and 2.23\% on the GlaS and MoNuSeg datasets, respectively.

Overall, the results across multiple datasets and segmentation tasks highlight the effectiveness of our proposed approach compared to a broad range of SOTA methods. Our MambaCAFU model demonstrates robustness while maintaining moderate complexity and computational efficiency, achieving an optimal balance between performance and resource requirements.

\subsection{Qualitative Results}

\begin{figure*}[htbp]
\setlength\tabcolsep{1pt}
\renewcommand{\arraystretch}{1.2} 
\begin{tabularx}{\linewidth}{XXXXXXXX}

\parbox[c][2.3cm][c]{2.3cm}{\includegraphics[width=2.3cm,height=2.3cm]{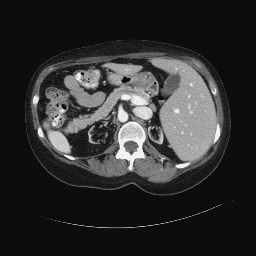}\vspace{1pt}} &
\parbox[c][2.3cm][c]{2.3cm}{\includegraphics[width=2.3cm,height=2.3cm]{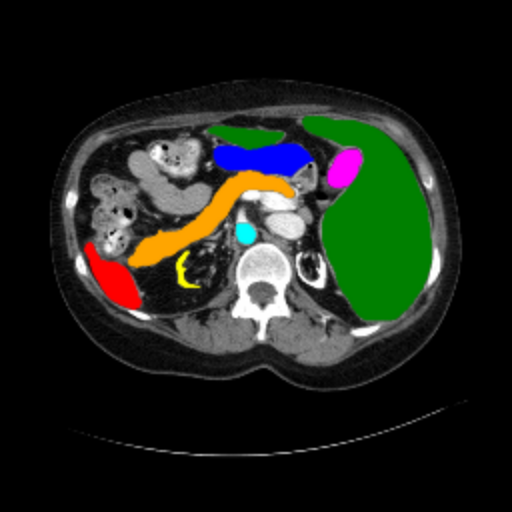}\vspace{1pt}} &
\parbox[c][2.3cm][c]{2.3cm}{\includegraphics[width=2.3cm,height=2.3cm]{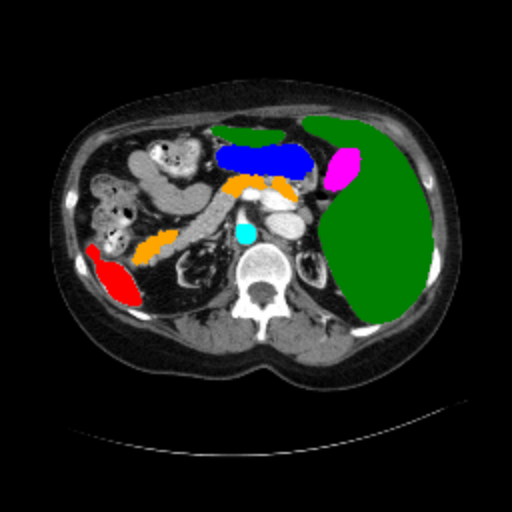}\vspace{1pt}} &
\parbox[c][2.3cm][c]{2.3cm}{\includegraphics[width=2.3cm,height=2.3cm]{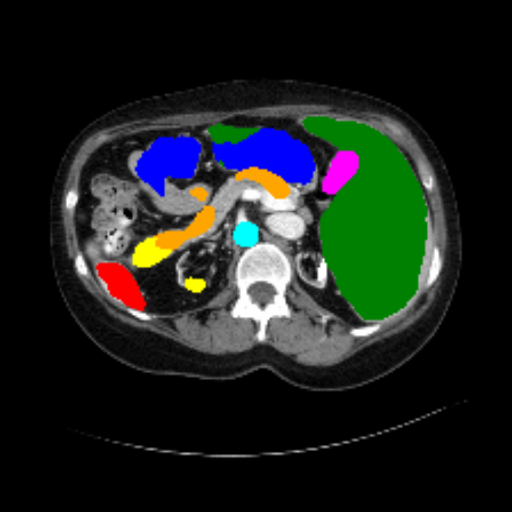}\vspace{1pt}} &
\parbox[c][2.3cm][c]{2.3cm}{\includegraphics[width=2.3cm,height=2.3cm]{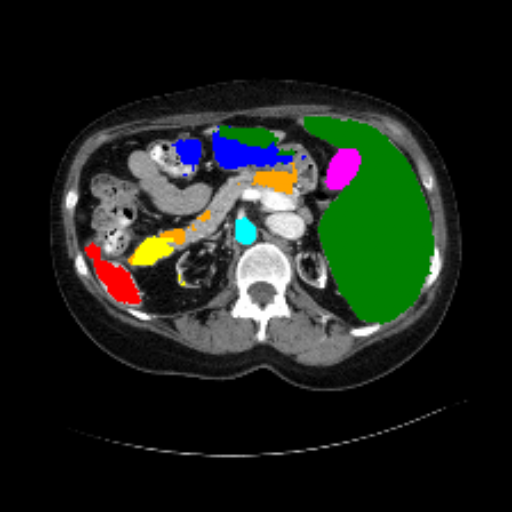}\vspace{1pt}} &
\parbox[c][2.3cm][c]{2.3cm}{\includegraphics[width=2.3cm,height=2.3cm]{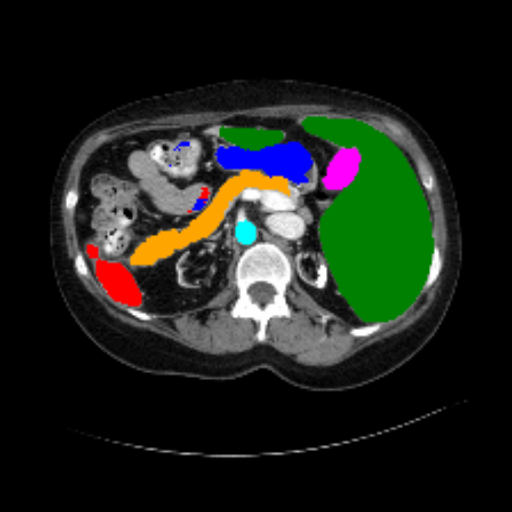}\vspace{1pt}} &
\parbox[c][2.3cm][c]{2.3cm}{\includegraphics[width=2.3cm,height=2.3cm]{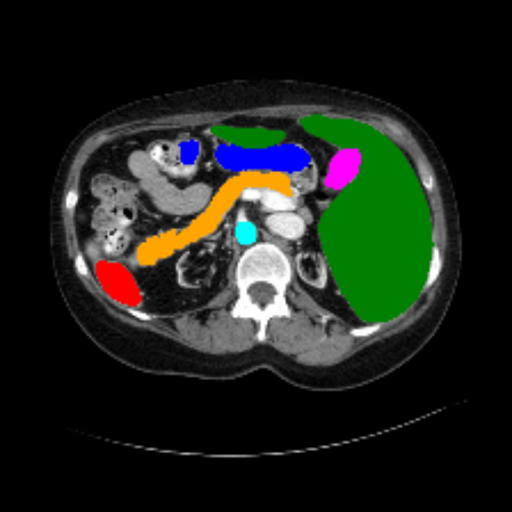}\vspace{1pt}} \\ \addlinespace[2pt]

\parbox[c][2.3cm][c]{2.3cm}{\includegraphics[width=2.3cm,height=2.3cm]{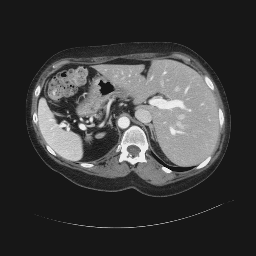}\vspace{1pt}} &
\parbox[c][2.3cm][c]{2.3cm}{\includegraphics[width=2.3cm,height=2.3cm]{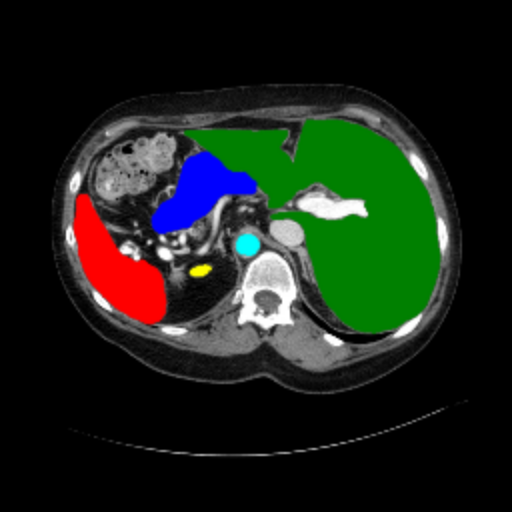}\vspace{1pt}} &
\parbox[c][2.3cm][c]{2.3cm}{\includegraphics[width=2.3cm,height=2.3cm]{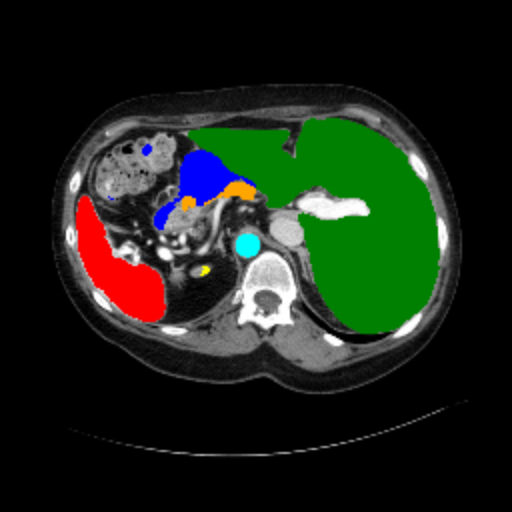}\vspace{1pt}} &
\parbox[c][2.3cm][c]{2.3cm}{\includegraphics[width=2.3cm,height=2.3cm]{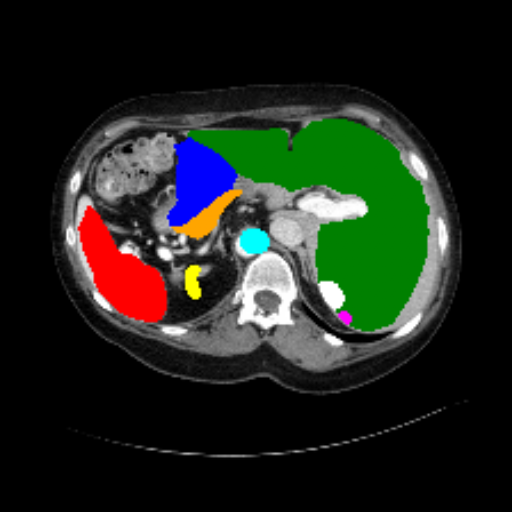}\vspace{1pt}} &
\parbox[c][2.3cm][c]{2.3cm}{\includegraphics[width=2.3cm,height=2.3cm]{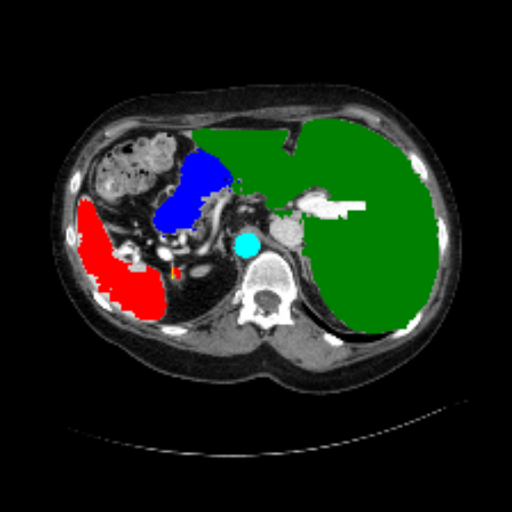}\vspace{1pt}} &
\parbox[c][2.3cm][c]{2.3cm}{\includegraphics[width=2.3cm,height=2.3cm]{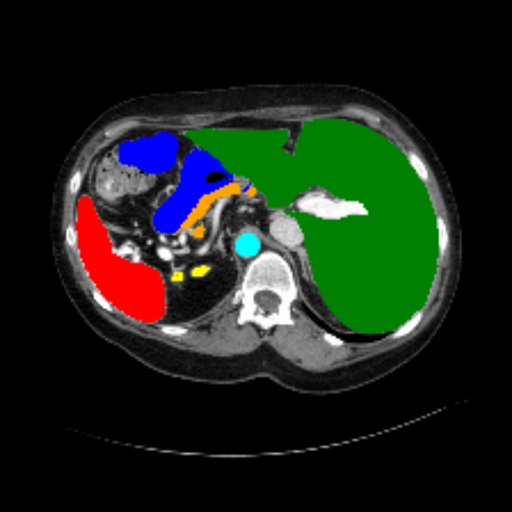}\vspace{1pt}} &
\parbox[c][2.3cm][c]{2.3cm}{\includegraphics[width=2.3cm,height=2.3cm]{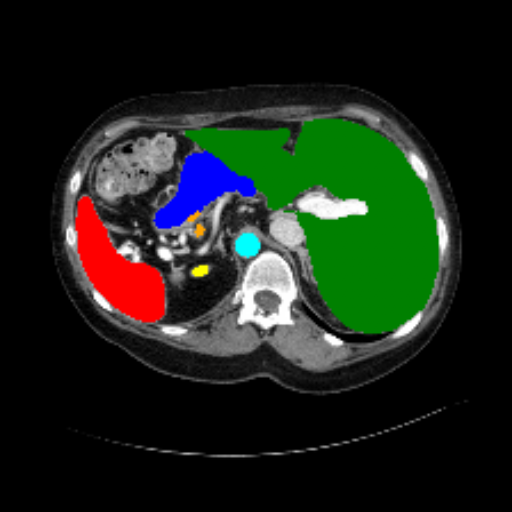}\vspace{1pt}} \\ \addlinespace[2pt]

\parbox[c][2.3cm][c]{2.3cm}{\includegraphics[width=2.3cm,height=2.3cm]{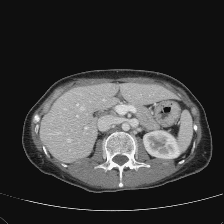}\vspace{1pt}} &
\parbox[c][2.3cm][c]{2.3cm}{\includegraphics[width=2.3cm,height=2.3cm]{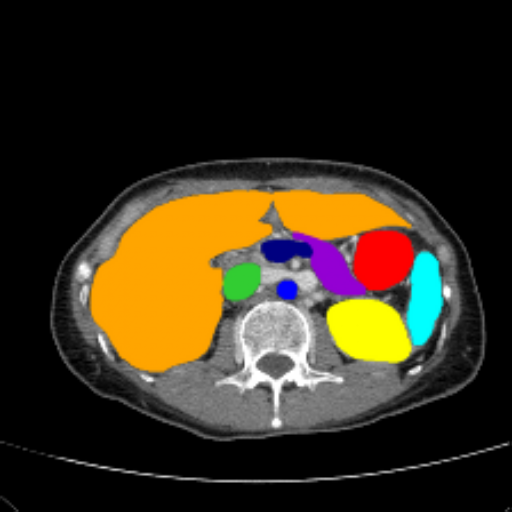}\vspace{1pt}} &
\parbox[c][2.3cm][c]{2.3cm}{\includegraphics[width=2.3cm,height=2.3cm]{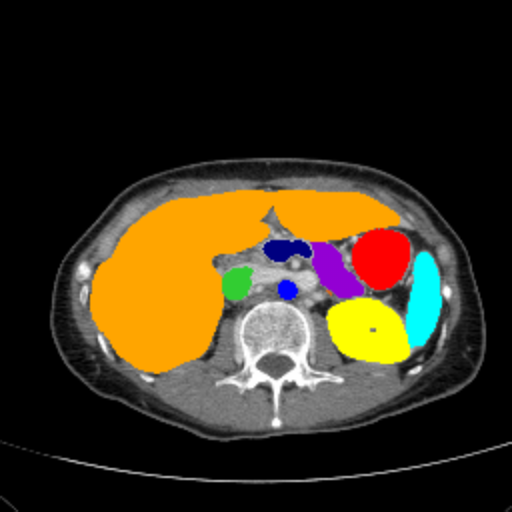}\vspace{1pt}} &
\parbox[c][2.3cm][c]{2.3cm}{\includegraphics[width=2.3cm,height=2.3cm]{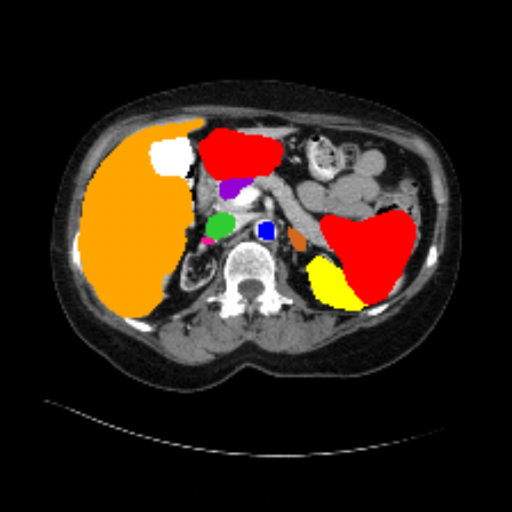}\vspace{1pt}} &
\parbox[c][2.3cm][c]{2.3cm}{\includegraphics[width=2.3cm,height=2.3cm]{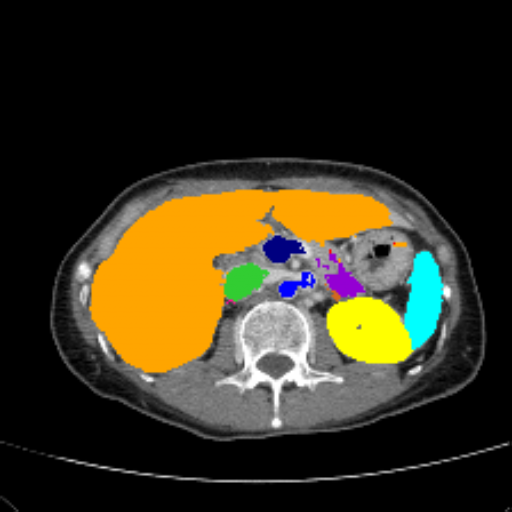}\vspace{1pt}} &
\parbox[c][2.3cm][c]{2.3cm}{\includegraphics[width=2.3cm,height=2.3cm]{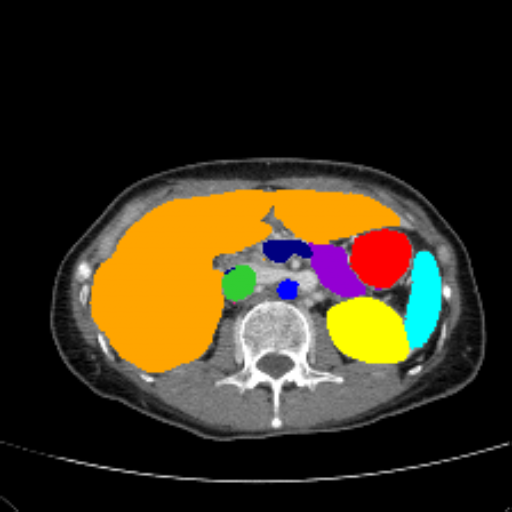}\vspace{1pt}} &
\parbox[c][2.3cm][c]{2.3cm}{\includegraphics[width=2.3cm,height=2.3cm]{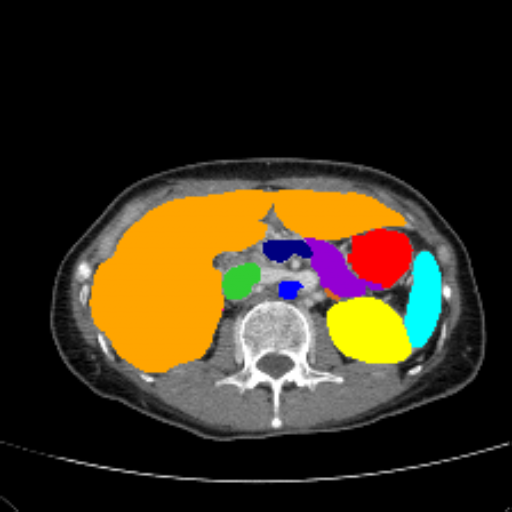}\vspace{1pt}} \\ \addlinespace[2pt]

\parbox[c][2.3cm][c]{2.3cm}{\includegraphics[width=2.3cm,height=2.3cm]{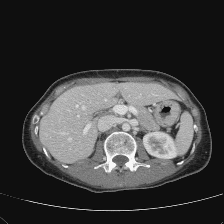}\vspace{1pt}} &
\parbox[c][2.3cm][c]{2.3cm}{\includegraphics[width=2.3cm,height=2.3cm]{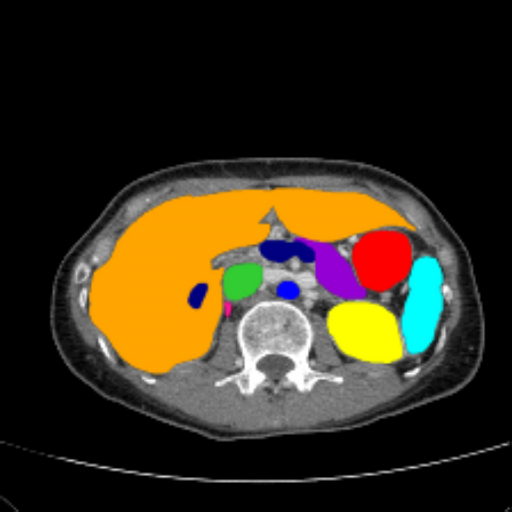}\vspace{1pt}} &
\parbox[c][2.3cm][c]{2.3cm}{\includegraphics[width=2.3cm,height=2.3cm]{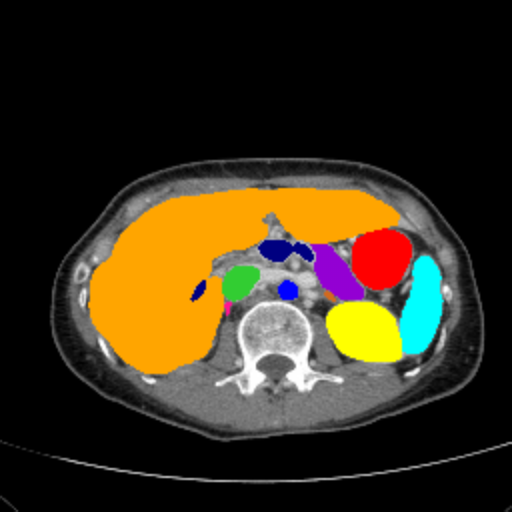}\vspace{1pt}} &
\parbox[c][2.3cm][c]{2.3cm}{\includegraphics[width=2.3cm,height=2.3cm]{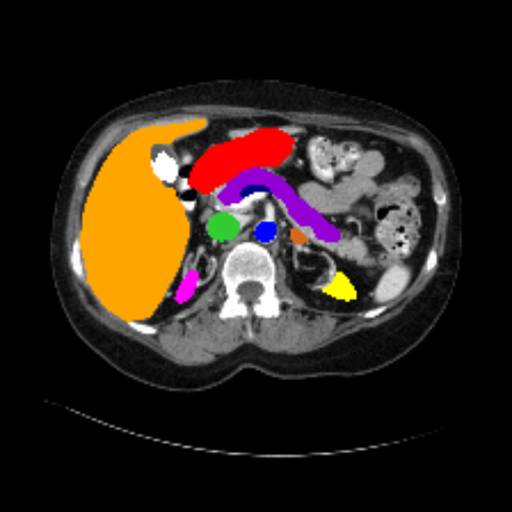}\vspace{1pt}} &
\parbox[c][2.3cm][c]{2.3cm}{\includegraphics[width=2.3cm,height=2.3cm]{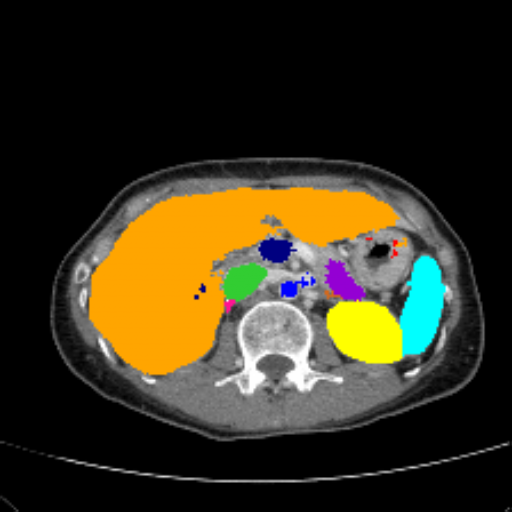}\vspace{1pt}} &
\parbox[c][2.3cm][c]{2.3cm}{\includegraphics[width=2.3cm,height=2.3cm]{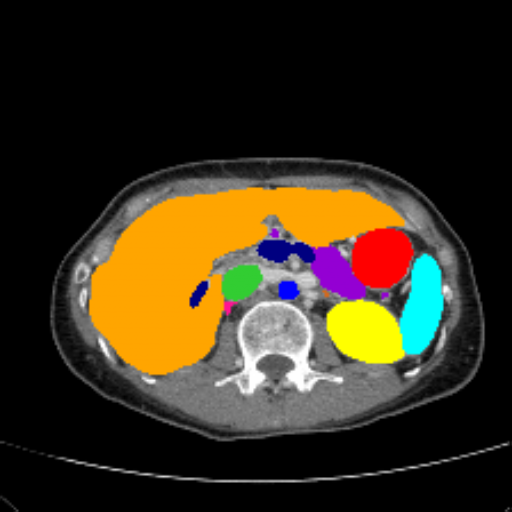}\vspace{1pt}} &
\parbox[c][2.3cm][c]{2.3cm}{\includegraphics[width=2.3cm,height=2.3cm]{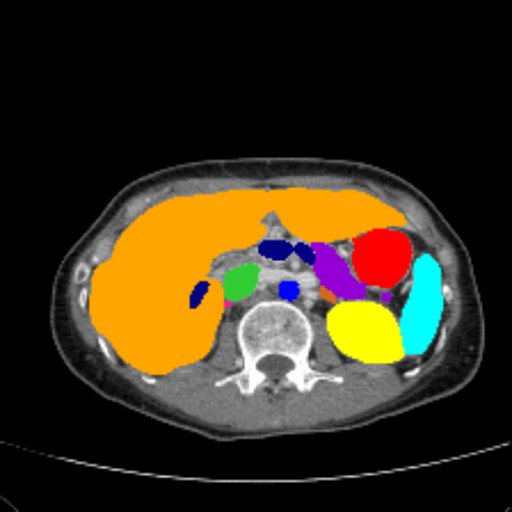}\vspace{1pt}} \\ \addlinespace[2pt]

\parbox[c][2.3cm][c]{2.3cm}{\includegraphics[width=2.3cm,height=2.3cm]{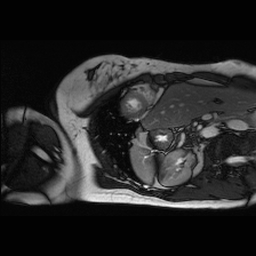}\vspace{1pt}} &
\parbox[c][2.3cm][c]{2.3cm}{\includegraphics[width=2.3cm,height=2.3cm]{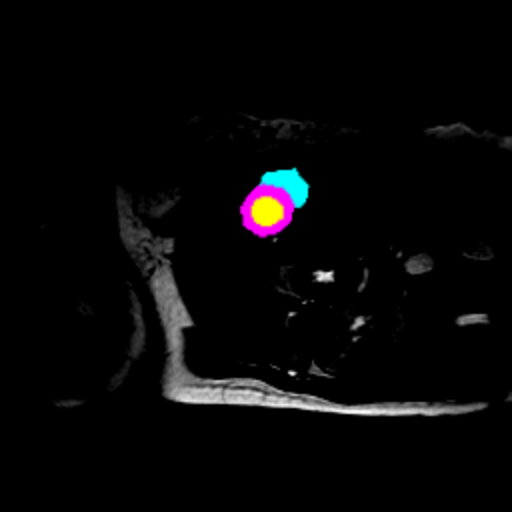}\vspace{1pt}} &
\parbox[c][2.3cm][c]{2.3cm}{\includegraphics[width=2.3cm,height=2.3cm]{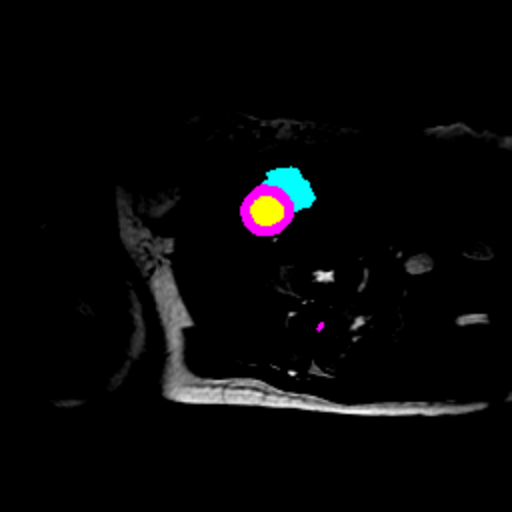}\vspace{1pt}} &
\parbox[c][2.3cm][c]{2.3cm}{\includegraphics[width=2.3cm,height=2.3cm]{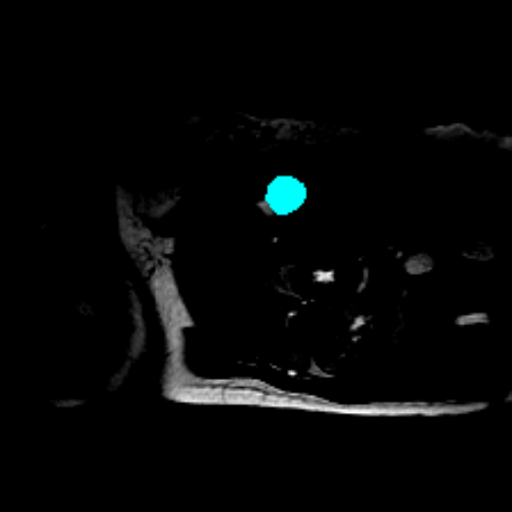}\vspace{1pt}} &
\parbox[c][2.3cm][c]{2.3cm}{\includegraphics[width=2.3cm,height=2.3cm]{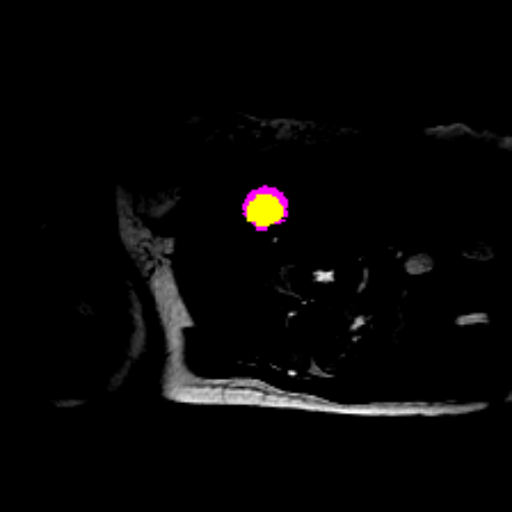}\vspace{1pt}} &
\parbox[c][2.3cm][c]{2.3cm}{\includegraphics[width=2.3cm,height=2.3cm]{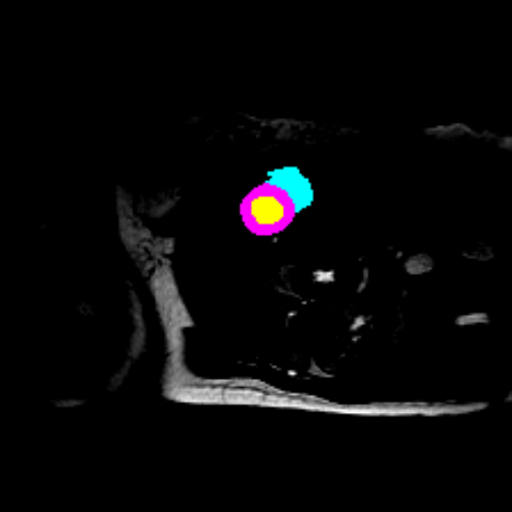}\vspace{1pt}} &
\parbox[c][2.3cm][c]{2.3cm}{\includegraphics[width=2.3cm,height=2.3cm]{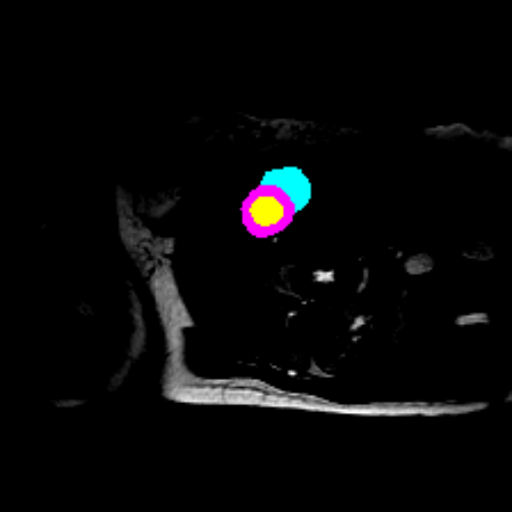}\vspace{1pt}} \\ \addlinespace[2pt]

\parbox[c][2.3cm][c]{2.3cm}{\includegraphics[width=2.3cm,height=2.3cm]{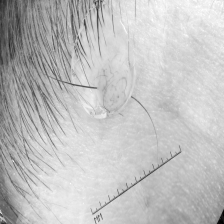}\vspace{1pt}} &
\parbox[c][2.3cm][c]{2.3cm}{\includegraphics[width=2.3cm,height=2.3cm]{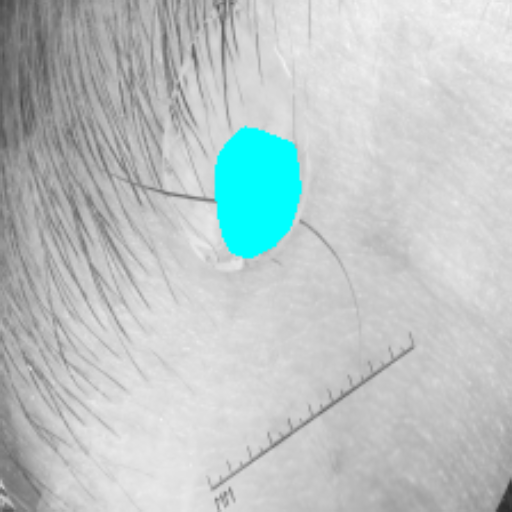}\vspace{1pt}} &
\parbox[c][2.3cm][c]{2.3cm}{\includegraphics[width=2.3cm,height=2.3cm]{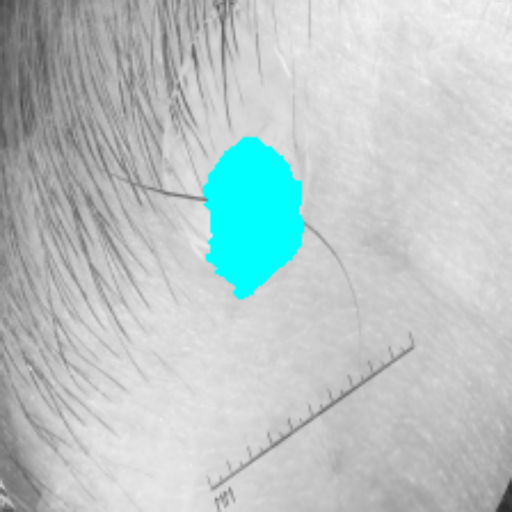}\vspace{1pt}} &
\parbox[c][2.3cm][c]{2.3cm}{\includegraphics[width=2.3cm,height=2.3cm]{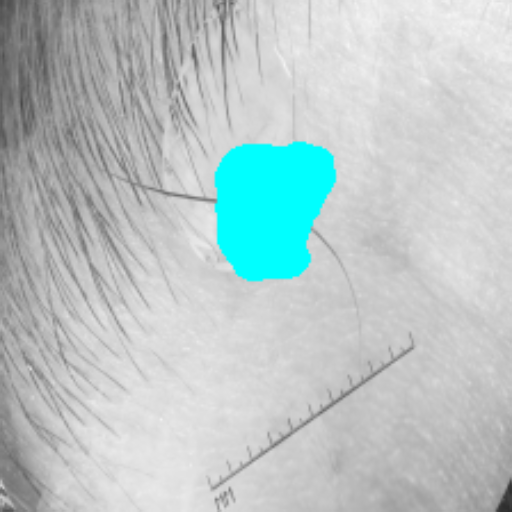}\vspace{1pt}} &
\parbox[c][2.3cm][c]{2.3cm}{\includegraphics[width=2.3cm,height=2.3cm]{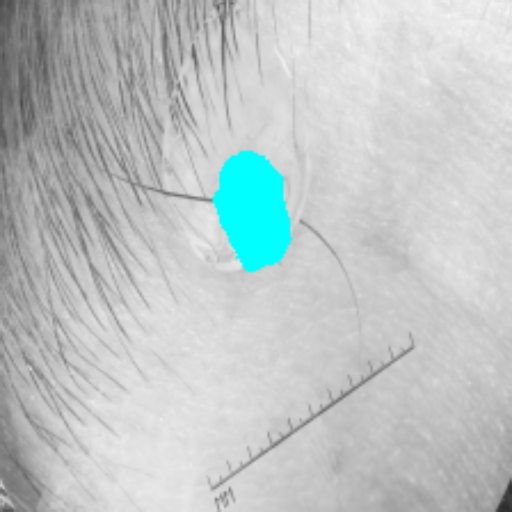}\vspace{1pt}} &
\parbox[c][2.3cm][c]{2.3cm}{\includegraphics[width=2.3cm,height=2.3cm]{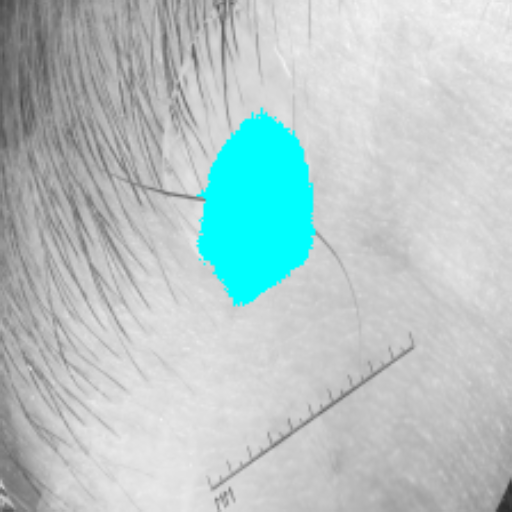}\vspace{1pt}} &
\parbox[c][2.3cm][c]{2.3cm}{\includegraphics[width=2.3cm,height=2.3cm]{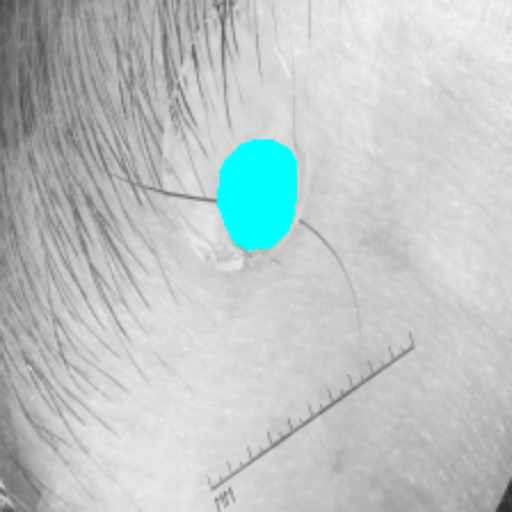}\vspace{1pt}} \\ \addlinespace[2pt]

\centering\bf{Slice} & \centering\bf{GT} & \centering\bf{Unet} &
\centering\bf{TransUnet} & \centering\bf{Mamba-Unet} &
\centering\bf{Swin-UMamba} & \centering\bf{MambaCAFU-V\textsubscript{1}} \\

\end{tabularx}
\caption{Visual comparison of segmentation examples from Synapse (first two examples), BTCV (3-4 examples), ACDC (5th example) and ISIC17 (last example). Columns: input slice, ground truth, Unet, TransUnet, Mamba-Unet, Swin-UMamba, and MambaCAFU-V\textsubscript{1}.}
\label{fig:compvisBTCV}
\end{figure*}

Figure \ref{fig:compvisBTCV} presents a qualitative comparison of segmentation performance on examples from the Synapse, BTCV, ACDC, and ISIC17 datasets. The first two examples (from Synapse) highlight variations in segmentation performance among U-Net, TransUnet, Mamba-Unet, and Swin-Unet. While Swin-Unet performs well in the first example, its performance decreases in the second example, particularly in segmenting the organ highlighted in blue. Additionally, it misclassifies background regions as the class highlighted in orange. In contrast, our approach demonstrates high robustness in segmenting all classes accurately and aligning well with the ground truth masks.

Across BTCV, ACDC, and ISIC17, the comparison methods exhibit varying performance depending on the task and class. For instance, Mamba-Unet struggles to segment multiple organs in BTCV, even misclassifying certain classes as others. In the ACDC dataset, both TransUnet and Mamba-Unet perform poorly in segmenting the three classes. In the last column (ISIC17), Swin-Unet oversegments the skin lesion compared to the ground truth. Unlike the comparison models, our approach consistently delivers effective segmentation across different tasks and datasets.

These qualitative findings further reinforce the quantitative results presented in the experimental section, demonstrating the effectiveness of our MambaCAFU model in handling binary and multi-class segmentation across diverse medical imaging modalities and tasks.


\subsection{Ablation study}
We conduct two ablation studies to assess the contribution of our proposed components. The first evaluates the impact of ResNet branch (ResB), Co-Attention Gate (CoAG), and MambaConv on the overall architecture. The second analyzes the effect of integrating our proposed blocks (CoASMamba, CoAMamba, and DoubleLCoA) into the encoder, bottleneck, and decoder, respectively.

In Table \ref{tab:abblocks}, Experiment 1 serves as the baseline, where ResB is removed, CoAG is replaced with simple concatenation, and MambaConv is substituted with standard convolutional blocks. Experiments 2 to 5 show that removing any component leads to performance drops on the Synapse dataset: 1.44\% (ResB), 1.22\% (CoAG), and 1.11\% (MambaConv), confirming their complementary roles. ResB improves localization, CoAG enhances multi-scale feature interaction, and MambaConv captures long-range dependencies.

Table \ref{tab:ablayers} examines the integration of our proposed blocks within each architecture stage. Compared to a baseline without them, MambaCAFU-V\textsubscript{1} achieves performance gains of 2.22\% and 2.36\% on Synapse and ISIC17, respectively. These results confirm the effectiveness of our blocks in improving feature representation and segmentation performance.

\section{Conclusion}
\label{S:7}
In this paper, we introduced MambaCAFU, a hybrid segmentation architecture that integrates CNNs, Transformers, and Mamba-based modules for medical image segmentation. Our model leverages pre-trained vision backbones with a novel Co-Attention Gate and Mamba-based Attention Fusion to enable efficient multi-scale feature learning. The encoder combines local and global representations, while the decoder employs multi-level co-attention and upsampling mechanisms for accurate segmentation across diverse medical imaging tasks. Extensive experiments and ablation studies confirm the effectiveness of each component, demonstrating that removing any key module degrades performance. By balancing computational efficiency with segmentation accuracy, MambaCAFU outperforms CNN-based, hybrid CNN-Transformer, and Mamba-based architectures, making it a promising solution for real-world clinical applications.


\end{document}